\documentclass{article}

\usepackage{arxiv}
\usepackage[utf8]{inputenc} 
\usepackage[T1]{fontenc}    
\usepackage{hyperref}       
\usepackage{url}            
\usepackage{booktabs}       
\usepackage{amsfonts}       
\usepackage{nicefrac}       
\usepackage{microtype}      
\usepackage{lipsum}

\makeatletter
\g@addto@macro{\UrlBreaks}{\UrlOrds}
\makeatother
\usepackage{hyperref}
\hypersetup{
  colorlinks = true,
  urlcolor   = [rgb]{0.094,0.094,0.498},
  linkcolor  = [rgb]{0,0.105,0.672}, 
  citecolor  = [rgb]{0,0.105,0.672}, 
}

\usepackage{graphicx}
\usepackage{caption}
\usepackage{subcaption}

\usepackage{mathtools}
\usepackage{textcomp}
\usepackage{gensymb}
\usepackage[gen]{eurosym}

\usepackage{makecell}
\usepackage{multirow}
\usepackage{amssymb}
\usepackage{pifont}
\newcommand{\cmark}{\ding{51}}%
\newcommand{\xmark}{\ding{55}}%

\newcommand{\UAB}{Universitat Autònoma de Barcelona}

\newcommand{\CVC}{Computer Vision Center}

\title{Identity Document and banknote \\ security forensics: a survey }

\author{
  Albert B.~Centeno \\
  \CVC \\
  \UAB \\
  \texttt{aberenguel@cvc.uab.es} \\
   \And
 Oriol R.~Terrades \\
  \CVC \\
  \UAB \\
  \texttt{oriolrt@cvc.uab.es} \\
   \And
 Josep L.~Canet \\
  \CVC \\
  \UAB \\
  \texttt{josep@cvc.uab.es} \\
    \And
 Cristina C.~Morales \\
  Senior Director of Research\\
  Mitek Systems, Inc.\\
  \texttt{ccanero@miteksystems.com} \\
}

\begin{document}
\maketitle

\begin{abstract}
Counterfeiting and piracy are a form of theft that has been steadily growing in recent years. Banknotes and identity documents are two common objects of counterfeiting. Aiming to detect these counterfeits, the present survey covers a wide range of anti-counterfeiting security features, categorizing them into three components: security substrate, security inks and security printing. respectively. From the computer vision perspective, we present works in the literature covering these three categories. Other topics, such as history of counterfeiting, effects on society and document experts, counterfeiter types of attacks, trends among others are covered. Therefore, from non-experienced to professionals in security documents, can be introduced or deepen its knowledge in anti-counterfeiting measures. 
\end{abstract} 


\section{Introduction}

Counterfeit is the action of make an exact imitation of something valuable with the intention to deceive or defraud. Usually counterfeit products are produced for dishonest or illegal purposes, with the intent to take advantage of the superior value of the imitated product. 

Counterfeiting goods is an important source of income for organized criminal groups. At the stage of distribution of counterfeit goods, fraudulent retail licences enable the infiltration of the legitimate supply chain. Generally, in order to run their illicit businesses, counterfeiters establish companies and bank accounts using fraudulent identity documents (ID) or under the name of a front person, and regularly make use of bogus invoices. Counterfeiters purchase or rent vehicles using fake documents. Number plates of cars belonging to criminal groups are registered abroad or under a fake identity. Fraudulent documents are widely used to facilitate the transportation, distribution and sale of counterfeit goods. For the purpose of importation, counterfeiters provide false shipping documents, such as bills of lading, to conceal the content of containers of packages and the origin of shipments. They often use false invoices issued for imported goods in declarations to customs. This practice is also used to undervalue their imported products. 

Counterfeit detection has traditionally been a task for law enforcement agencies, see section. EUROPOL and INTERPOL central offices are combating document and banknote counterfeiting~\cite{interpol,europol}. They have destined millions of euros in funds to provides technical databases, forensic support, training and operational assistance to its member countries. Despite all these efforts counterfeit detection remains an open issue. It makes no difference how many security barriers the experts place in their way, the criminal competitors are never far behind. As a consequence, the production of counterfeit money and IDs is on the rise. 

There are many different strategies used to fake an ID like the alteration of a real passport, impersonation of the legitimate owner or printing false information on a stolen blank real paper, to cite some. Making a fake passport is easy, making a good fake passport is very, very hard. Probably there are few criminal organizations in the world which can produce a counterfeit visa or passport good enough to fool professional passport control. Same reasoning applies to banknotes counterfeit production. It is important to understand which are the types of attacks a counterfeiter can produce.

Document security experts are a key element to detect high-quality counterfeits. They usually have received an intensive training and have available forensic equipment to evaluate the authenticity of the suspicious documents. However most of the population has not receive any training into identify a counterfeit banknote or identity document. This lack of knowledge causes that fraudsters continue using counterfeited security documents causing a harmful effect to the society. Therefore there exists a need for tools to aid the reviewers of security documents. 

In this survey of identity document and banknote security forensics we focus in the anti-counterfeit security measures which can be solved automatically by computer vision algorithm. The purpose of using computer vision algorithms is to automate the authentication process of security documents by means of affordable tools, such a smartphone, that the general population could use. A question that could arise to the reader of this article is: What is the difference between the presented survey and others surveys in the literature? We compare in Table~\ref{tab:survey_comparison} the contribution of this survey with respect to other state-of-art surveys in counterfeit detection. We order the comparison by year and topics it covers each survey. 

\begin{table}[h]
 \caption{Survery comparison. Identity/Banknote docs: survey done for IDs or banknotes. Counterfeit history: history of counterfeit. Effects Society: causality between the counterfeit and the effects in society. Document experts: figure of a document expert. Security Substrate/Ink/Printing: what are the anti-counterfeit features for these categories. Type of attacks: types of attacks done by the counterfeiters. Digital Tampering: digital watermarking approaches. Datasets: datasets of the presented approaches. Approaches: state-of-art works in counterfeit. Systems and apps.: state-of-art works contain systems and applications for mobile devices. Trends: general direction guidelines. *: Present work.}
  \centering
  \begin{tabular}{lcccccccccc}
    \toprule
    Features     & \cite{rudolf_1995} & \cite{singh2013survey} & \cite{mahajan2014survey} & \cite{mann2015comparative} & \cite{Chambers2015} & \cite{chakraborty2016review} & \cite{lee2017survey} & \cite{hassan2017survey} & \cite{Upadhyaya_survey} & *  \\
    \midrule
    Year        & 1995      & 2013      & 2014      & 2015      & 2015       & 2016      & 2017  & 2017      & 2018     & 2019 \\
    Identity docs       & \cmark &          &        &        &        &        &        & \cmark &        & \cmark \\
    Banknote docs       & \cmark &          & \cmark & \cmark & \cmark & \cmark & \cmark &        & \cmark & \cmark \\
    Counterfeit History &        &          &        &        & \cmark &        &        &        &        & \cmark \\
    Effects Society     &        &          &        &        &        &        &        &        &        & \cmark \\
    Document Experts    &        &          &        &        &        &        &        &        &        & \cmark \\
    Security Substrate  &        &          &        & \cmark & \cmark &        &        &        & \cmark & \cmark \\
    Security Ink        & \cmark &          &        & \cmark & \cmark &        &        &        & \cmark & \cmark \\
    Security Printing   & \cmark &          &        & \cmark & \cmark &        & \cmark &        & \cmark & \cmark \\
    Types of attacks    &        &          &        &        &        &        &        &        &        & \cmark \\
    Digital tampering   &        &  \cmark  &        &        & \cmark &        &        & \cmark &        & \cmark \\
    Datasets            &        &          &        &        &        &        & \cmark &        &        & \cmark \\
    Approaches          &        &  \cmark  & \cmark &        & \cmark & \cmark & \cmark & \cmark & \cmark & \cmark \\
    Systems and apps.   &        &          & \cmark &        & \cmark & \cmark & \cmark & \cmark & \cmark & \cmark \\
    Trends              &        &          &        &        &        &        &        &        &        & \cmark \\
    \bottomrule
  \end{tabular}
  \label{tab:survey_comparison}
\end{table}

The work in~\cite{rudolf_1995} presents a survey of security features containing optically invariable and variable devices to measure its practical value for document security. It does not compare algorithms or approaches and it is for both  banknote and identity documents. The main feature of this work is that it orders the security features by degrees of order and degrees of security. Being the degrees of order the size in mm need to inspect each one of the security features. More than 20 years has past since this survey was done, so it is not up to date with the new security features. In~\cite{singh2013survey} an overview and comparison of digital watermarking techniques is presented. Although this survey does not fit entirely to the study of anti-counterfeits features in security documents, it does introduce the concept of securing documents by introducing unique watermarking codes. The survey in~\cite{mahajan2014survey} centers in counterfeit paper currency recognition and detection. They detect fake banknotes, but some of the approaches presented, centers the counterfeit detection in the recognition of the banknotes instead of the anti-counterfeit security features. If the banknote is not recognized by the classifier then it may be considered as a fake banknote. The work in~\cite{mann2015comparative} does a comparative study on security features of banknotes. It does not compare approaches, but show graphically different examples of banknotes what are the security features and where are located. The thesis in digital currency forensics in~\cite{Chambers2015}, contains a review for the security anti-counterfeit measures in banknotes, it contains some of the sections presented at the present survey. The fact that is a thesis and not a survey, explains why it needs a better comparison of the approaches presented and a clearer structure of the security features works explained.  The survey in~\cite{chakraborty2016review} for currency note authentication techniques lacks of a better comparison between the different approaches and expand the literature compared. In~\cite{lee2017survey} a complete and easy to read survey on banknote recognition methods is presented. They focus in the recognition and counterfeit detection, with a large literature. They also include many comparisons and even does an study of the datasets available in recognition and counterfeit banknote detection. The inclusion of the recognition process of the banknotes makes that the counterfeit part is not as complete as it should. Our work also has a better explanation of each of the security measure present in security documents. In a more recent survey for conterfeit currency detection techniques in~\cite{Upadhyaya_survey} explains what are most of the anti-counterfeiting techniques and which Rupee banknote denominations includes them. Most of the works compared are centered on the Rupee. This work lacks of a comparison between approaches presented. Finally, the only survey in identity documents is presented in~\cite{hassan2017survey}. Different approaches are presented and explained what could be the improvements and weaknesses of each method, but needs a better comparison between those methods and also to explain the dataset context to understand the experimental results. 

From the surveys presented in Table~\ref{tab:survey_comparison}, we believe not enough importance has been given to digital tampering. It has been demonstrated that humans are easily fooled by tampered images~\cite{schetinger2018humans, kasra2018seeing, Nightingale2017}. When no original images are given for comparison, people have an extremely limited ability to detect and localize image tampering. The availability of internet and the access to efficient and high quality image editing software for the majority of the population, creates a window for counterfeiters to be introduced in the forging world or to perfect their counterfeiting techniques. Nowadays the research of algorithms for detect tampering is more needed than ever. Research community has put a lot of effort into develop algorithm to automatically discover tampered images. We divide this section in two depending on the availability of the original image and its access. If the original image has never been we propose several tampering detection algorithms, on the other hand if its possible to be in contact with the manufacturer which grants access to the original image, it is possible to use watermarking techniques.

Another section which is not frequently mentioned in the survey literature is the dataset section. Here we found one of the main drawbacks for comparing anti-counterfeit algorithms. There is no public available datasets for counterfeit detection in IDs and banknotes. This leads to every researcher to build their own private datasets where it will extract some results that in most cases nobody would be able to reproduce. Also the difficulty of create these datasets to gather both genuine and counterfeit samples makes that each private dataset which generally contains few samples. Typically, this small subset of samples difficultly will represent the big variety of cases in the open-world scenarios. 

Similarly, usually does not exist a dedicated section in the literature based on systems and mobile applications for document counterfeit detection. At this survey we create a section focusing into the state-of-art of complete systems or smartphone applications, because we believe one of the main purposes is to provide a wide range of population a set of tools to check for counterfeits in an accessible, affordable and comprehensible fashion.

Our survey categorizes the anti-counterfeit security measures into three components: security substrate, security inks and security printing. Security substrate corresponds to the security measures associated with the materials used to produce the security documents and the embedded features introduces in it. Security inks can be classified as the types of inks that react when detects an attempt to fraudulently alter a document, or inks used verify the document authenticity. Usually manufacturers use a combination of both types of security inks, alongside other security features. Lastly, security printing corresponds to the printing of variable information between security documents of the same type. In IDs, security printing usually corresponds with the bio-data information of the owner. This work explains in detail the different components of each of the three categories of security measures and presents different algorithmic approaches in the literature for these measures.


The remainder of the paper is organized as follows. Section~\ref{section:brief_history} briefly explains the history of counterfeit and how it has been evolving to this day. Section~\ref{section:effects_society_and_document_experts} introduces the effects that nowadays is causing counterfeit in the society and the document security expert figure. Next section~\ref{section:datasets} presents the datasets created in the field of security documents and its availability. Section~\ref{section:anti-counterfeit-measures} describes different types of anti-counterfeit security features available to detect fraud security documents. This section is divided into security substrate, security inks and security printing. 
Information about conterfeiter types of attacks is described in~\ref{section:types_of_attacks_vulnerabilities}. It also presents the tampering types and tampering detection approaches if the original genuine document is not accessible, and watermarking techniques otherwise.
On section~\ref{section:approaches-methodologies} we present approaches and algorithms in the literature for banknote and security document counterfeit. 
Section~\ref{section:systemsAndapplications} presents the main systems and architecture applications created to date. Section~\ref{section:trends} does an analysis of the trends for anti-counterfeiting technologies for the next years. Finally, in section~\ref{section:conclusion} we draw the conclusions and future work needs to be done to continue updating this survey and further expand the knowledge about counterfeiting in the future.

\section{A brief history of counterfeit detection}
\label{section:brief_history}

Counterfeit is as old as the alphabet or the money itself, sometimes referred too as the "2\textsuperscript{nd} oldest profession" in the world~\cite{wikipedia_counterfeit_money}. In the ancient world, it was not unusual for the workers at the forge to duplicate coins by using gold plated bronze and not pure gold. Augustus Caesar and other rulers of the day were quick to see the implications and imposed heavy penalties, often death. Shaving the coin edges was a common practice to produce counterfeit coinage. Laws against counterfeit can be traced to the years $80$BC when the Romans established a permanent court to try cases involving forgeries of all sorts, including currency counterfeit~\cite{Barlow_romans} or falsification of documents that transferred land to heirs~\cite{Koppenhaver2007}. 

Chinese started carrying folding money during the Tang Dynasty (A.D. 618-907), mostly in the form of privately issued bills of credit or exchange notes~\cite{pickering1844history}. Wood from mulberry trees was used to make the money. To control access to the paper, guards were stationed around mulberry forests, punishing thieves entering the forests to death. Since then, the crime of counterfeit money has been practiced in every country where writing existed and paper was used for financial transactions. Europe took around $500$ years more to start using paper bills, where the practice began to catch in the 17th century. English couple Thomas and Anne Rogers were convicted for counterfeiting 40 pieces of silver. Thomas was hanged, drawn and quartered while Anne was burnt alive. Forms of punishment were considered acts of treason against state or Crown, rather than simple crime. In 1739, similarly in America, Benjamin Franklin intentionally misspelled the word ‘Pennsylvania’ on his bills to catch forgers who corrected the error~\cite{Moneymakers}. In the late 18th and early 19th centuries, Irish immigrants to London were associated with the spending of counterfeit money~\cite{Irish_counterfeit}. 

Particularly the production of counterfeit money has been used by nations as a means of warfare, to overflow the enemy's economy with useless fake banknotes, so that the real value of the money plummets. Great Britain use this strategy during the American Revolutionary War to reduce the value of the Continental Dollar. During the American Civil War, counterfeited Confederate States dollar was mass produce by private interests on the Union side. Thanks to the access to modern printing technology, the imitations were often equal of even superior quality compared with the Confederate money. In the 1920s Hungary was engaged in a plot to purchase 10 million fake Francs as a move to avenge their territorial losses in World War I. Unsuccessfully during World War II, the Nazis attempted to collapse the Allies economy (Operation Bernhard)~\cite{Operation_Bernhard}. Jewish artists in the Sachsenhausen concentration camp were forced to forge British pounds and American dollars. The outstanding quality of the counterfeit money made almost impossible to distinguish between the real and fake bills. However the Nazis could not carry out the planned aerial drops with the counterfeit money over Britain and America.  

Today the most sophisticated counterfeit bill ever produced and undetectable even to currency experts are the "Superdollars", because of their high quality, and likeness to the real US dollar. The origin of this banknotes is unclear, where North Korea, Russia or even the CIA has been accused~\cite{counterfeit_superdollar,counterfeit_superdollar_CIA}. 

Not as well documented as the history of counterfeit money, notorious forgers of identity documents have also existed since the old days. After the end of the World War II, many officials and high-ranking Nazis forged identity documents to flee from Germany. Like Adolf Eichmann, referred as the "architect of the Holocaust", escaped to Argentina using a "laissez-passer" issued by the International Red Cross under a fraudulently identity~\cite{Adolf_Eichmann}. Alexander Viktorovich Solonik, a hitman and a Russian gangster in the early 1990s, lived in Greece using a fake passport issued in the consulate in Moscow~\cite{Alexander_Solonik}. Also famous was the arrest of Kim Jong-nam, the son of North Korean dictator Kim Jong-il, who was detained by Japanese immigration official travelling with a forged Dominican Republic passport~\cite{Kim_Jong_nam}. One of the greatest counterfeiters of the 20th century was Adolfo Kaminsky~\cite{Adolfo_Kaminsky}. A former member of the French Resistance during the World War II, forged papers to save the lifes of $14.000$ Jews. Afterwards he continued forging papers for various groups during 30 years trough different wars~\cite{kaminsky2016adolfo}. 

Last decades the control of identity documents and banknotes were exclusively controlled by document experts. Usually country border controls or banks did not had at their disposal automatic software to validate security document, which made the authentication prone to human errors. The apparition of computer vision and machine learning in the 50s-60s decades, has helped to develop new algorithms to automatically detect counterfeits documents. With machine learning it is even possible to render new text in someone's handwriting, producing novel images of handwriting that look hand-made to casual observers, even when printed on paper~\cite{HainesTOG2016}. Nowadays counterfeiters are using AI and machine learning to make better fakes~\cite{counterfeiters_ai}. At the same time researchers and authorities are developing new methods using AI to spot them.

Government authorities and counterfeiters have been playing a game of cat-and-mouse, as soon as new security features are added to the security documents, criminals try to copy them. Today, unlike a millennium ago under the rule of Emperor Augustus, fraudsters don’t need to fight lions in the Roman stadium if they are caught, however severe forms of punishment exists differentiated by country~\cite{Penalties_by_country_for_creating_counterfeit_money}. Banks and government authorities need to have strong lines of defense against fraudsters of security documents. If they find themselves the weakest link, they can guarantee fraudsters will attack.

\section{Effects of forgery in society and document experts}
\label{section:effects_society_and_document_experts}

Counterfeit objects produced for criminal activity not only causes potential harm to the health and safety of the citizens, it also affects legitimate economies, contributing to reduced revenues for the affected businesses, decreases sales volume and job losses. According to a 2013 report, the OECD estimated counterfeit goods accounted for $2.5\%$ of global trade \cite{OECD_2016_counterfeit}. Currently, trade of counterfeit and pirated goods represents $\$1.7$ trillion per year and is expected to grow to $\$2.8$ trillion and cost $5.4$ million jobs by 2022~\cite{forbes_2018_counterfeit}. In 2018, counterfeit was the largest criminal enterprise in the world, more than drugs and human trafficking~\cite{forbes_2018_counterfeit}. This cost represents all the existing types of counterfeit objects, next we focus in IDs and banknotes counterfeits. 

From the different types of counterfeit document fraud continues to play a key enabling role in trafficking of counterfeit goods. Counterfeiting goods is an important source of income for organized criminal groups. At the stage of distribution of counterfeit goods, fraudulent retail licences enable the infiltration of the legitimate supply chain. Generally, in order to run their illicit businesses, counterfeiters establish companies and bank accounts using fraudulent identity documents (ID) or under the name of a front person, and regularly make use of bogus invoices. Counterfeiters purchase or rent vehicles using fake documents. Number plates of cars belonging to criminal groups are registered abroad or under a fake identity. Fraudulent documents are widely used to facilitate the transportation, distribution and sale of counterfeit goods. For the purpose of importation, counterfeiters provide false shipping documents, such as bills of lading, to conceal the content of containers of packages and the origin of shipments. They often use false invoices issued for imported goods in declarations to customs. This practice is also used to undervalue their imported products. 

Banknote counterfeiting is another illegal lucrative business for the counterfeiters. Through recorded history currency has been used as a medium of exchange for goods and services. Paper notes, coins and electronic currency are the general accepted form of trade. Governments of each country using central banks are the responsible to issue money and circulate it within an economy. Currency only holds its value as long as users have confidence in its authenticity to represent goods. A common threat for any economy is the quantity of counterfeit money which is being used in the actual market. Higher than previous years, in the first half year of 2015, $454$K counterfeit Euro banknotes were withdrawn from circulation, being $86\%$ of the counterfeits from \euro{}20 and $\euro{}50$ banknotes,\cite{intro_euro_counterfeit}.

The \emph{modus operandi} in counterfeiting and piracy have partially changed over the past few years and are expected to evolve further in the future. To be able to carry out such a variety of activities, criminal groups need stable access to resources. However, this does not seem to pose any major difficulty for criminals. Widely available and affordable information and communication tools increasingly facilitate their activities.


A natural consequence of the existence of counterfeit objects is the creation of the document expert figure. Document experts play a key role to reduce the cost and other potential dangers counterfeiting causes in the society. Document security experts are highly skilled in ascertaining document security threats, identifying material and technology weaknesses, and developing innovative solutions to safeguard against counterfeiting and alteration.

The nature of document counterfeits is such that the initial encounter with a document requiring authentication is rarely within a specialized document forensics laboratory~\cite{siegel2012encyclopedia}. A passport may first be viewed by immigration or customs officers, currency by a shop assistant or a bank clerk, identity documents by a transport authority officer, etc. At this first step is where most of the counterfeits go unnoticed, just a small percentage are detected if the inspection is done by an a person without the proper training. After some assessment, if a security feature integrated in the document seems altered, the document is given to a specialist examiner. 

The examiner looks for combinations of significant similarities and combinations of significant differences between the questioned document and an exemplar document. If the examiner finds combinations of significant similarities between the questioned and the exemplar, the examiner may conclude it is dealing with a genuine document. Otherwise, if significant differences exist between the compared documents he may determine is a case of counterfeit. Although this principle of comparison seems simple and sound, the reality is far from simple. The terms "significant similarities" and "significant differences" are subjective~\cite{eckert1996introduction}. The examiner must ultimately decide what is significant and what is not. These decisions come from the examiner's knowledge and understanding of class characteristics, individual characteristics, and all the environmental facts that can affect the security document, such as the wear and tear, dirt, lightning conditions on how the document is checked, etc. The authenticity response is highly dependent with the document examiner knowledge. The knowledge is acquired through intensive supervised training and much practical experience. The economic effort to train a document  expert is high and usually there is a shortage of people who can effectively do this task.

The research of algorithms to automatize this chain of processing of inspection is to make the authentication less prone to human errors. Nowadays, algorithms to detect counterfeit documents are still far from the accuracy of document experts and their examination laboratories. 60\% of fake documents can be detected through detection machines or algorithms for counterfeit detection while 80\% can be detected by document experts~\cite{mandridaketowards}. These data shows that most of the fraudulent documents have not yet been detected. Most researchers focus nowadays into the first level of the chain, where counterfeit detection is done by untrained personnel, and is possible to catch a bigger percentage of fraudulent documents.

\section{Datasets}
\label{section:datasets}

The evaluation of counterfeit detection algorithms is a constant challenge for researchers. Building a counterfeit dataset \emph{per se} represents a difficult task due the scarcity nature of counterfeit documents. Usually a counterfeit dataset contains a small percentage of counterfeits compared with their genuine counterpart. Counterfeit datasets are usually collected by documents experts, see section~\ref{section:effects_society_and_document_experts}. Training a document expert is expensive, hence generating a dataset generated by them represents a big economical effort. Most private companies in document security analysis can afford to invest in the generation of counterfeit dataset, however make these datasets public does not play in its own interests. On the other hand, even having the economical means and the predisposition of building a public dataset, is difficult to publish it as a benchmark for the research community. The copyright status of the security documents designs is carefully controlled by counterfeiting laws. In the case of the banknotes, this copyright status can vary from nation to nation~\cite{wikimedia_commons_copyright}. Some of the restrictions imposed at the banknotes, for instance the Euro banknotes, are harmful for evaluating computer vision algorithms. Euro banknotes are copyright of the European Central Bank. There are rules such as the word SPECIMEN must be printed diagonally across the reproduction, in a non-transparent colour contrasting with the dominant colour of the note. Also width and height of the word must represent at least 75\% and 15\% respectively of the document. Moreover the resolutions of the shared image must not exceed 72 dots per inch (dpi). These restrictions, established to prevent counterfeiters to fabricate imitations, also harms the use of algorithms which requires higher image resolutions to look for authenticity details. 

Despite all these obstacles, researchers have produced algorithms to solve the banknote counterfeit problem creating their own datasets. One work around to publish datasets of banknotes is to process the images with your algorithm and then share only the output feature data \cite{LohwegDataSet2012}. These features, extracted from genuine and forged banknote-like specimens, contains Wavelet Transformed image (variance, skewness, curtosis) information and entropy information. These features, corresponds to patches of 400x400 pixels of the banknote. The patches are digitized with an industrial camera, at close distance from the banknote to the lens. Due the close distance to the investigated banknote the resulting resolution is 600 dpi~\cite{Lohweg2012}. However this dataset does not allow to develop new computer vision algorithm development approaches. 

Most of the researchers works directly with the banknote images and do not share their datasets due the commented previous limitations. A recente survery analyzes 45 datasets for banknote recognition methods and only one is publicly available~\cite{lee2017survey}. The public available dataset corresponds to Jordan bills and coins acquired with a smartphone on different backgrounds~\cite{doush2017currency}. In the survery they also explore 16 counterfeit banknote detection datasets, used in 31 research publications, where none of them are public. At Table~\ref{tab:identity_counterfeit_dataset_availability} we present ID and banknote counterfeit datasets used at the current literature. We present the availability, the number of document samples, the number of classes and the percentage of counterfeits it contains. 

\begin{table}
 \caption{Datasets on ID and banknote counterfeit detection (Ref.: References, N/I: Not Informed, N/A: Not Available).}
  \centering
  \begin{tabular}{lcccccccc}
    \toprule
    Availability & B/ID & \#Images & \#Classes & $\mu$ & $\sigma$ & \%Counterfeit &  References \\
    \midrule
    N/A & B & 60K & 6 & 10K & 0 & - & \cite{jin2008hierarchical}  \\
    N/A & B & 99 & 1 & 99 & 0 & 29\% & \cite{yeh2011employing}  \\
    N/A & B & - & 7 & - & - & - & \cite{Bhavani2014}  \\
    N/A & B & 357 & 3 & 119 & 1 & - & \cite{chae2009study}  \\    
    N/A & B & 1K & 1 & 1K & 0 & 50\% & \cite{roy2015machine}  \\
    N/A & B & 900 & 6 & - & - & - & \cite{Glock2009}  \\
    N/A & B & 264 & - & - & - & - & \cite{Lohweg2012}  \\
    N/A & B & 66 & 1 & 66 & 0 & 50\% & \cite{Lohweg2013}  \\
    N/A & B & 2.75K & 7 & - & - & - & \cite{bruna2013}  \\
    N/A & B & 82 & 2 & - & - & 48\% & \cite{dewaele2016forensic}  \\
    N/A & B & 200 & 1 & - & - & 50\% & \cite{Chang2007}  \\
    \midrule
    N/A & ID & 50 & 22 & - & - & - & \cite{kwon2007recognition}  \\
    A & B/ID & 14.7K & 62 & 236 & 314 & 25\% & \cite{berenguel2016banknote,aberenguel:evaluation,aberenguel:econterfeit}  \\
    \bottomrule
  \end{tabular}
  \label{tab:identity_counterfeit_dataset_availability}
\end{table}

The dataset collected in~\cite{Chang2007} is collected on the street including new and worn out banknotes. Local defects of tears, stains, graffiti or holes for fitness classification and counterfeit detection are manually labeled in~\cite{jin2008hierarchical}. In~\cite{roy2015machine} the regions of the rupees banknotes are scanned using four different 4 different light sources, which are UV light, Co-axial light and Flood light with two different magnitude and gain configurations. Then the authors uses the source light that suits best the security feature they want to analyze. A dataset for intaglio textures authentication is build in~\cite{Glock2009}. The dataset is divided in textures printed with Intaglio printing process and by offset printing. The dataset is also classified into high-quality and medium-quality prints. They scan at 1200 dpi and convert to grayscale 6 different classes of textures. Both counterfeit and genuine banknotes have been acquired under several environmental lighting conditions, with different illuminants and brightness in~\cite{bruna2013}. The authors also introduces in the dataset misalignment with slightly translated and or rotated banknotes. In~\cite{dewaele2016forensic} the genuine banknotes are scanned at 1200 dpi, to print them later with an inkjet and laser printer. Genuine press printed banknotes, inkjet and laser ones are acquired in a later stage with a smarphone. 

As in banknotes, identity document designs are copyrighted in most countries. Additionally, identity documents contains Personally identifiable information (PII). PII is any data that could potentially identify a specific individual. Information which, when disclosed, could result in harm to the individual whose privacy has been breached. PII can be exploited by criminals to stalk or steal the identity of a person, or to aid in the planning of criminal acts. Laws and regulations from each country ensures their citizens are protected against fraudulent use of the PII contained at their identification documents. Countries differs in the approach to data protection. For instance, in Europe, privacy and data protection appear as fundamental freedoms, contained in the General Data Protection Regulation (GDPR). All these restrictions and protection laws makes very difficult for the research community to share a public dataset to be used as a benchmark for counterfeit ID detection. The authors in~\cite{arlazarov2018midv} have created a public video dataset for ID analysis and recognition using smartphones. Although the dataset does not contain counterfeit documents, the 500 video streams can be used for anomaly detection or as pretraining for a counterfeit classification model. In~\cite{kwon2007recognition} they created a passport dataset for MRZ and bio-data comparison. In~\cite{berenguel2016banknote,aberenguel:evaluation,aberenguel:econterfeit} the authors create a counterfeit dataset of banknotes and IDs. The images are acquired as a normal user could do in a non-controlled environment at close undetermined distance. The images presents different kinds of perspective distortions and some of them can have occluded areas. They use six different acquisition devices: two digital cameras, one tablet and three mobile cameras. The illumination conditions are also variable due the images are indoor and outdoor. The dataset is available under demand to the authors, except the ID images due PII restrictions. The IDs and banknotes samples represent 3.1K and 11.5K respectively. The dataset is bias to contain \EUR{} and IDs with a total of 9.7K samples. The counterfeits are only produced for the \EUR{} bankotes and IDs with a total of 3.6K samples, representing a 37.7\% for this set, and a 25\% of the total dataset.

\section{Anti-counterfeit measures}
\label{section:anti-counterfeit-measures}

Banknotes and identity documents contain specific security features for protection against counterfeit and fraud. Each year more and more security features are included in their designs in order to ward off potential counterfeiter, fraudsters and impostors. A single feature hardly can provide the level of security needed for this type of documents, usually a combination of different anti-counterfeit measures is used to secure a document.  

Security patterns are specially designed with distinctive characteristics in the hope that people can easily recognize them. Three easy-to-follow methods can distinguish genuine banknotes or documents from counterfeit ones: feel, look and tilt. The first method correspond to touch the material of the document and check the surface does not present anomalies. The second method is to observe the document when hold against the light and compare it with a known genuine document. Last method, tilting indicates the security measures printed with special ink, optically variable when viewed at different angles. Moreover it is possible to check for more advances security features with specialized equipment. The security features are designed to resist deterioration for reasonable wear and tear and robust to forgery. 

The security features are designed to resist deterioration for reasonable wear and tear and robust to forgery. Security features against forgery can be categorize in three types: Substrate materials used for the fabrication of the document, the type of ink and the printing design, see Table~\ref{tab:security_features_types}.

\begin{table}
 \caption{Security features types classified by fabrication materials, ink used and printing process or design. Not all existing security types are included.}
  \centering
  \begin{tabular}{lll}
    \toprule
    Substrate     & Ink     & Printing \\
    \midrule
    Complex substrate recipe & Complex ink recipe   & Offset Lithographic   \\
    Windowed security thread & Colour-shifting ink  & Intaglio printing     \\
    Security fibres          & Ultra violet ink     & Personalization        \\
    Watermark                & Infrared ink         &                       \\
    See through windows      &                      &                       \\
    \bottomrule
  \end{tabular}
  \label{tab:security_features_types}
\end{table}

\subsection{Security Substrate}
\label{subsection:security_substrate}

Paper results from compression of different plant fibres. The substrate of security paper is manufactured for one particular application and for one particular contractor only; hence, it is not commercially available for the general public. Generally, the substrate is made of paper, almost always from cotton fibres for strength and durability. The security paper is usually provided with chemical reactants, watermarks, fibers, planchettes, and threads to add individuality and protect against counterfeiting. In later stages various types of mechanical perforation and laser perforation may be put to use to further enhance the security level. The majority of security paper manufacturers prefer the production of security documents with paper due its lower cost. Gaining popularity is the use polymer security documents made from BOPP, which stands for Biaxially-oriented Polypropylene~\cite{prime2010australia}. The polymer fabricated documents are longer lasting, harder to destroy, waterproof, have better dirt resistance, and can be recycled when taken out of circulation decreasing the environmental impact. However, polymer documents can not be easily folded and can be permanently damaged if exposed to a heat of around $100\degree$ C. All security features from paper can be incorporated in polymer documents and allows to include new security features which can not be applied to paper. For instance, the inclusion of a small transparent windows (few milimeters in size) as a security feature, also name see through windows, is difficult to reproduce using common counterfeiting techniques~\cite{izdebska2015printing}. Polymer documents usually incorporate \emph{Optically Variable Device} (OVD) as a security feature and are very hard to counterfeit simply because many of its unique security features cannot be reproduced by scanning and photocopying them. Brightness reflected by the substrate composition can be also used as a security feature~\cite{yeh2011employing, Bhavani2014, Desai2014ImplementationOM}.  However, brightness will also be affected by the inks and printing. Table~\ref{tab:summary_substrate_section2} presents a summary of the presented methods at
this section.

\begin{figure}
     \centering
     \begin{subfigure}[b]{0.3\textwidth}
         \centering
         \includegraphics[width=\textwidth,height=3cm]{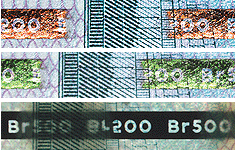}
         \caption{500 Ruble security thread.}
         \label{fig:500_ruble_security_thread}
     \end{subfigure}
     \hfill
     \begin{subfigure}[b]{0.3\textwidth}
         \centering
         \includegraphics[width=\textwidth,height=3cm]{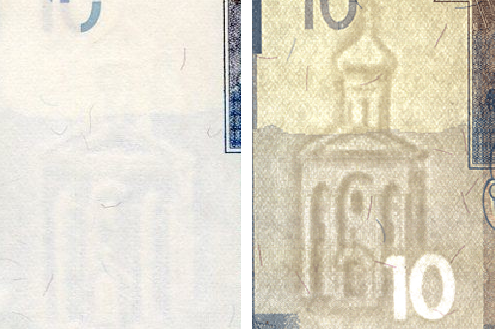}
         \caption{10 Ruble watermark.}
         \label{fig:10_ruble_watermark}
     \end{subfigure}
     \hfill
     \begin{subfigure}[b]{0.3\textwidth}
         \centering
         \includegraphics[width=\textwidth,height=3cm]{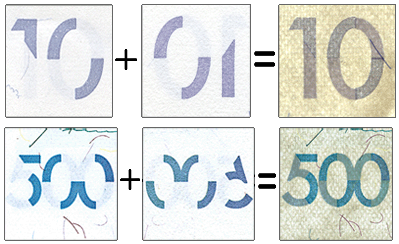}
         \caption{10-500 Ruble see-through.}
         \label{fig:see-through-smartphone}
     \end{subfigure}
    \caption{Example of anti-counterfeit features based on the substrate of Belarusian rubles. Belarusian currency is not copyrighted. (a) Metallic thread of a windowed type. When viewed against the light the security thread is seen as solid dark. (b) Watermark combined with half-tone in the form of the fragment of the main image 10 Ruble obverse, filigree (light) of the "10" denomination banknote. (c) When held up to the light, averse and reverse fragments are filled with colors of the opposite side, forming the complete banknote denomination number.}
    \label{fig:anti-counterfeit-substrate-based}
\end{figure}

\subsubsection{Substrate embedded security features}
\label{subsubsection:security_substrate_embedded}

Opacity is an intrinsic property to the paper substrate. It describes the amount of light which is transmitted through the paper. A complete opaque object is one which allow no light to pass through it. Cellulose fibers, the primary component of paper, are transparent, but stacking them and creating a web structure with them diffuses the light through the sheet and increases the paper opacity. Paper opacity determines the extent to which printing on a particular side of paper will be visible from the reverse side, named \emph{see-through}, see Figure~\ref{fig:see-through-smartphone}. Manufacturers exploit this property to embed latent security components between the layers of the substrate. This also applies to polymer substrate where latent images or security components are hidden from normal view. Opacity also affects to the printed inks on the substrate, determining the level of transparency of the security document. Opaque pigments will block light to pass through and transparent pigments will allow varying amounts of light to pass trough the substrate, revealing the reverse side background printing of a sheet of paper. 

\emph{Watermarks} is a very well-known and reliable security feature for protecting documents against counterfeiting, see Figure~\ref{fig:10_ruble_watermark}. The fabrication process of a watermark is as follows: A cylinder covered by a wire mesh embossed with the watermark design rotates in a vat containing cotton pulp. The cylinder mould process is the preferred way to embed a watermark for banknotes and IDs. The suspension of cotton fibres is agitated in the vat and the wire mesh retains the fibrous material in the hollow areas. The variations of fibre density forms the image of the watermark. The variation of fibre density produces areas with different paper thickness. Varying thickness of paper produces different shades of lightness/darkness when holding the document up to the light, or shining one through the paper. When a genuinely watermarked paper is held to the light, the thicker areas of the paper appear dark, and when placed beneath a light the dark areas appear lighter. The watermarks are used for displaying portraits and motifs. Given the high level of recognition of watermarks around the world, even tiny defects in the portrait or the motif are detected instantly. Different works include the watermark analysis~\cite{Chang2007,prasanthi2015indian}. While the new polymer banknotes produced do not have watermarks, an analogue security feature named \emph{Shadow image} is produced by setting an image into one of the polymer substrata during manufacture. When observed in transmitting light, the Australia’s coat of arms shadow image appears in the first polymer series of the Australian \$50 note.

\emph{Security threads} are threads of natural or synthetic material placed in the paper during manufacture. Incorporated at the beginning of the paper production process, similarly to the security fibre, is embedded between the layers of the substrate, except it is placed in a regular position. Different variations of security threads has been developed. "Morse code" thread has solid and translucent sections. It is possible to read Morse code characters at the broken line created by the solid region when held to the light. Similarly "Microprinted" thread, shows microprinted writing on the translucent section. Usually, in banknotes the writing correspond to the initials of the issuing authority. The "contoured" thread is a wide thread that has one straight side and one wavy side, with the wavy side pointing either to the left or right on the document. "Windowed" threads, is the latest development in the use of security threads, more difficult to forge as threads are woven in and out of the note surface. Viewed normally it looks like it is appearing and disappearing at regular intervals at the surface of the paper, see Figure~\ref{fig:500_ruble_security_thread}. When held up to a light source it appears as a continuous line, although slightly broader. These variations can be combined to add further complexity to the security thread. Security threads are widely used in banknote and ID papers to deter counterfeiting, being reliable features against photocopying. 

\begin{figure}
     \centering
     \begin{subfigure}[b]{0.47\textwidth}
         \centering
         \includegraphics[width=\textwidth,height=3cm]{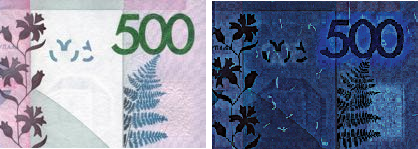}
         \caption{500 Ruble security fluorescencing fibers yellow-green.}
         \label{fig:500_ruble_security_fibers}
     \end{subfigure}
     \hfill
     \begin{subfigure}[b]{0.47\textwidth}
         \centering
         \includegraphics[width=\textwidth,height=3cm]{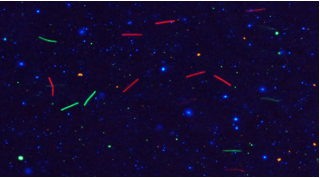}
         \caption{Planchettes and fibers from specimen security paper.}
         \label{fig:planchettes_and_fibers}
     \end{subfigure}
    \caption{Example of anti-counterfeit features embedded on the substrate visible under ultraviolet light. (a) Security fluorescencing fibers of yellow-green color in a 500 Ruble banknote. (b) Planchettes and fibers in a specimen security paper. Security fibers are red-green, meanwhile planchettes can be green-orange-blue in this specific case. In (a) and (b) they fibers and planchettes are scattered randomly in the paper substrate.}
    \label{fig:see-through-image}
\end{figure}

Woven into the layers of the substrate, \emph{coloured and fluorescent fibres} are embedded within the paper during its manufacturing. They appear as thin elements scattered all through the paper. These fibres are not visible under normal light, but under ultraviolet light (UV) the threads glow, see Figure~\ref{fig:500_ruble_security_fibers}. They represent an effective feature to protect any document at a cost-effective price and are present in most banknotes and passports. Authors in~\cite{Chang2007} uses spectral analysis to the reflected signal of fluorescent fibres, because counterfeits most likely will not contain hidden fluorescent fibers.

Likewise security fibres or threads, it is possible to implant small printed pieces of metal in the substrate, also named \emph{planchettes}. They held the same properties as when held up to a light source or exposed to UV light only, see Figure~\ref{fig:planchettes_and_fibers}. Planchettess are minute disks, metallic or transparent, ranging from about 1 mm to 5 mm in diameter. Microprinted text or symbols can be added to the planchettes.     

Usually counterfeiters include imitation security fibres in their document replicas, being this one a low security measure, however it is not obvious to the general public. The same applies to the security thread or planchettes, depending on their complexity.

\subsubsection{Spectrography}
\label{subsubsection:security_substrate_Spectrography}

Another security measure is the analysis of the chemical composition substrate. Some of the previous substrate security features can only be inspected with specialized devices capable of producing different light waves to the security document, like ultraviolet light. These devices are based on \emph{Spectroscopy}. Spectroscopy is the science related with the measuring and investigation of spectra produced when matter interacts with or emits electromagnetic radiation. For the analysis, a device separates separates incoming light waves reflected from the document substrate into a frequency spectrum for its analysis. Using a spectrography microscope it is possible to analyze different security components hidden to the naked eye, like the security fibers or embedded motifs made with security inks, see section~\ref{subsection:security_inks} for more details. Most of this forensic analysis can be done using infrared and UV spectrum. A procedure based on the analysis of several areas of euro banknotes using microscope ATR-infrared spectroscopy is proposed in~\cite{vila2006development}.    

\emph{Mössbauer spectroscopy} is a non-destructive chemical analysis which probes very small changes in the energy levels of an atomic nucleus in response to its environment. Using this technique it is possible to determine the atomic composition of the pigments used in the substrate. The concentration of pigments in the printer ink used and the specificity of their Mössbauer spectra can be used to identify fakes and forgeries~\cite{rusanov2002mossbauer}. Same authors showed that Mössbauerr and X-ray fluorescence studies revealed that a significant number of banknotes are printed using pigments which contain considerable amounts of iron~\cite{rusanov2009mossbauer}.

Similarly to Mössbauer, \emph{Raman spectroscopy} is a powerful method for material identification, capable of recognize different substances and their structural modifications. Any differences in the chemical ink composition or the paper should appear in their Raman spectra as a presence or an absence of particular peaks and their distribution in spectrum. An important drawback of this method is that Raman spectra has a weak effect in comparison with luminiscence security features. The stronger quantum effect of the luminiscence intensity can overlap the Raman spectra and mask spectral information.

\subsubsection{Fingerprinting paper surface}
\label{subsubsection:security_substrate_fingerprinting}

Low-cost physically unclonable functions (PUFs) are functions to create intrinsic random physical features to identify objects. A mathematical analogy of PUFs are the hash functions. The PUFs applied by the physical system follows the next properties: easy to evaluate and get a response, its outputs look like a random function, always return a unique response for the same request and the response has to be unpredictable even for an attacker with physical access to the object. 

PUFs as been receiving increasing attention in both research community and industry for counterfeit detection. Paper surface formed by overlapped and inter-twisted wood fiber forms an inherent unique 3D structure. The imperfections of a surface paper sheet caused by the manufacturing process can be used to uniquely identify the paper~\cite{buchanan2005forgery}. It is extremely unlikely that two document surfaces created with the same raw materials will be identical, although they will present similarities. Paper texture lead to unique maps of surface norm which can be transformed to a unique digital representation of the paper, named \emph{paper fingerprint}. Hence, paper fingerprints are PUFs applied to the substrate of the security document. 

Paper fingerprints can be effectively extracted with commodity scanners, scanning the paper surface from four different angles and then construct a 3D model, which later can be condensed into a feature vector~\cite{clarkson2009fingerprinting}. Paperspeckle is another approach to extract the paper speckle patterns at microscopic level \cite{sharma2011paperspeckle}. The authors use a microscope (with a 10-200X zoom) joined with an inbuilt LED source as the light which falls on a paper sheet. They microscope is then connected to a mobile device. A binary fingerprint is built with the randomly mixed dark and bright regions formed by the scattered light. They also demonstrate how this method produces a repeatable fingerprint even if the paper surface is damaged due to crumpling, printing or scribbling, soaking in water or aging with time. Following the same line of work, other works carries out an study of how high resolution photos of paper surface acquired distantly using industrial acquisition devices have good authentication performance, whereas the extension into using built-in cameras of mobile phones has acceptable performance at a higher computational cost~\cite{voloshynovskiy2012towards,diephuis2013physical,diephuis2014framework}. The industrial acquisition device (resolution of 5Mp) builds a micro-structure database of fingerprints under a controlled lighting environment. Later the verification can be done with a handheld camera (resolution of 2MP without optical magnification) in a different external environment. The drawback of this approach it critically depends on excellent lighting and acquisition conditions. Smartphone cameras have not a substantial success in obtaining consistent appearance images due to the uncontrolled nature of the ambient light. To solve this drawback, it is possible to use multiple camera-captured images at different viewpoints to estimate the paper surface~\cite{wong2017counterfeit}. Exploiting the camera flashlight the authors create a semi-controlled lighting conditions. Although authors use different smartphones to acquire a square-shaped paper dataset, needs further study on how it could performs with ID documents or banknotes exposed to day to day degradation caused by the normal used.  

It is also possible to exploit for PUF authentication purposes the embedded paper features instead of just focus in the substrate texture. The generation of \emph{spontaneous bubbles} in a polymer it is being used commercially for anti-counterfeiting purposes~\cite{Prooftag}. They use a transparent polymer material that generates bubbles at complete random when manufactured. The bubbles positions, sizes and shapes constitutes a unique fingerprint impossible to replicate which is sensible to small variations. The use of randomly distributed \emph{visible fibers} or \emph{color dots} on surfaces can be also used to provide uniqueness~\cite{Kinde,Prooftag}. A comparison study of some of the mentioned previous PUFs to gain a better understanding of the factors affecting the performance under mobile imaging~\cite{wong2015study}. They claim that due the uncontrollable light sources, and limits in camera resolution and focusing capability, the patch image intensity maps have a bad performance for pixel-domain correlation detector. They have also found that the density of foreground objects at the paper textures have a strong impact on the authentication performance. 

A novel paper fingerprinting technique is proposed by the authors in~\cite{toreini2017texture}. They propose to fingerprint the paper sheet based on its \emph{texture patterns} instead of features on the surface as performed in previous works. An analysis of the translucent patterns revealed when a light source shines through the paper. The fingerprinted patterns represent the random interleaved wooden particles inherent to the manufacturing process of the texture paper. They report zero error rates for the collected databases and to be robust against various distortions such as crumpling, scribbling, soaking and heating. The authors also demonstrate that the embedded paper texture provides a more reliable source for fingerprinting that feature on the surface. A drawback to this method is that the light needs to go through the document to be able to analyze the translucent patterns. They also use a camera-based acquisition device able to acquire images at high resolution and able to capture photos in a macro mode from a short distance (minimum 1 cm focus). A study should be done with smartphone cameras in a similar settings. 

\begin{table}
 \caption{ Summary of research works of anti-counterfeiting substrate features for
banknotes.}
  \centering
  \begin{tabular}{llll}
    \toprule
    Type                 &   Feature            &    Method             & References \\
    \midrule
    Substrate            &   Brightness information    &  luminance histogram      & \cite{yeh2011employing, Bhavani2014, Desai2014ImplementationOM} \\
    \cline{2-4}
    \multirow{2}{*}{Spectrography}            &   Iron concentration    &  X-rays and Mösbauer      & \cite{rusanov2009mossbauer} \\
                        &   IR patterns    &  Fourier transform infrared      & \cite{vila2006development} \\
    \midrule
    \multirow{2}{*}{Embedded}             &   Watermarks, Security Threads    & binary segmentation   & \cite{prasanthi2015indian}       \\
                        &   Watermarks, Security fibers    & spectral analysis   & \cite{Chang2007}       \\
    \midrule
    \multirow{5}{*}{PUF}                 &   paper substrate texture    & norm map                               & \cite{clarkson2009fingerprinting} \\
                     &   paper substrate texture    & texture microstructure                         & \cite{voloshynovskiy2012towards,diephuis2013physical,diephuis2014framework,sharma2011paperspeckle, wong2017counterfeit} \\
                     &   embbeded paper features       & 2D/3D bubble shape and location               & \cite{Prooftag} \\
                     &   embbeded paper features      & fibers, coloured dots shape and location      & \cite{Prooftag,Kinde} \\
                     &   embbeded paper features      & translucent patterns              & \cite{toreini2017texture} \\
    \bottomrule
  \end{tabular}
  \label{tab:summary_substrate_section2}
\end{table}

Authenticating security documents using PUFs seems extremely robust in theory, especially through values stored in a database. From the banknote and IDs authentication point of view, using PUFs requires further studies of its resilience against tampering of the surface paper, like scratching, folding, crumpling, and on the reliability of the physical structure over their lifespan. The fingerprint could change over time due the damage that naturally and inevitably occurs as a result of normal wear or aging. Making irrelevant the initial stored database hash fingerprint for the authentication.

\subsection{Security Inks}
\label{subsection:security_inks}

All document security features can be categorized by their level of user concealment, named \emph{overt}, \emph{semi-covert} and \emph{covert}. Overt security features indicates is directly perceptible by one or more of the human senses without recourse to external devices. Overt security features provide an instant verification, a predictable (repeatable) response based on the angle of light and could be either persistent or disappearing. They include elements such as holograms, watermarks and security threads. Oppositely, covert means not directly perceptible by the unaided human senses and detectable only through the use of purpose-built tools or professional laboratory equipment. Covert features include ultraviolet images, hidden text or images and other hidden items embedded into a document. On the other hand, semi-covert stands for not directly perceptible by the human senses but detectable by those senses through the use of non-professional external devices. Semi-covert security features requires foreknowledge and action, providing an instant verification and being either reversible or irreversible. Examples of semi-covert features is the micro printing or chemical markers. At this section we focus in several types of security inks categorized by their concealment level, see Table~\ref{tab:security_inks}. Chemical and physical analysis of inks on questioned documents provides valuable information regarding their authenticity.

\begin{table}
\caption{Security ink types classified by levels of concealment. Covert means a hidden ink invisible to the naked eye. Overt has the opposite meaning of covert and allows an inspector to verify the ink with a glance. Semi-covert stands for inks invisible to the human senses but detectable through the use of non-professional external devices. }
  \centering
  \begin{tabular}{lll}
    \toprule
    Overt ink    & Semi-Covert ink  & Covert ink    \\
    \midrule
    Optical Variable      & Thermochromic       & Biometric              \\
    Holographic           & Metameric           & Invisible UV           \\
    Iridescent            & Photochromic        & Invisible IR           \\
                          & Phosporescent       & Magnetic               \\
                          & Fugitive            & Machine-readable       \\
                          & Reactive            &                        \\
    \bottomrule
  \end{tabular}
  \label{tab:security_inks}
\end{table}

\begin{figure}
     \centering
     \begin{subfigure}[b]{0.47\textwidth}
         \centering
         \includegraphics[width=\textwidth, height=3cm]{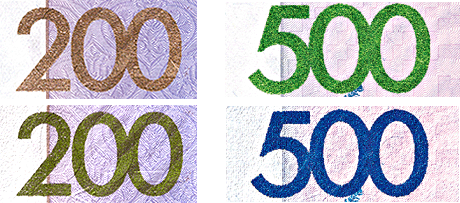}
         \caption{200-500 Ruble OVI.}
         \label{fig:200_500_ruble_ovi}
     \end{subfigure}
     \hfill
     \begin{subfigure}[b]{0.47\textwidth}
         \centering
         \includegraphics[width=\textwidth, height=3cm]{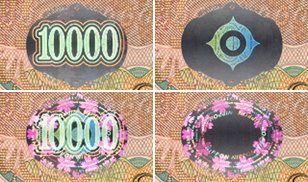}
         \caption{10K Japanese Yen hologram}
         \label{fig:hologram}
     \end{subfigure}

     \begin{subfigure}[b]{0.47\textwidth}
         \centering
         \includegraphics[width=\textwidth, height=3cm]{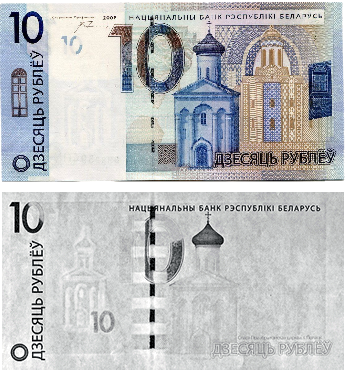}
         \caption{10 Ruble invisible ink.}
         \label{fig:10_Ruble_invisible_ink}
     \end{subfigure}
     \hfill
     \begin{subfigure}[b]{0.47\textwidth}
         \centering
         \includegraphics[width=\textwidth, height=3cm]{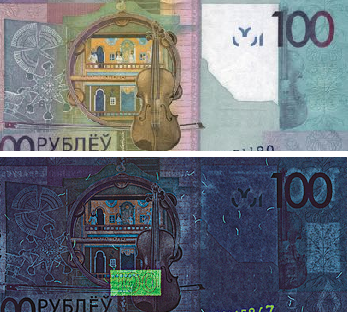}
         \caption{100 Ruble fluorescence ink.}
         \label{fig:100_ruble_fluorescence_ink}
     \end{subfigure}

    \caption{Example of anti-counterfeit ink features of Belarusian and Japanese banknotes. Japanese banknotes are exempt from copyright protection. (a) Optically Variable Ink (OVI). The denominations numbers change color when being viewed at different angles. (b) Holographic ink. When seen from different angles the denomination characters or different designs appears. (c) Invisible ink. Ruble banknote seen on the Infra-red (IR) band of the spectrum. (d) Fluorescence ink. Security fibres and a green rectangle with the denomination appear under UV light.}
    \label{fig:example-different-security-inks}
\end{figure}

Starting with overt inks types, \emph{Optical Variable ink} (OVI) is a type of ink composed by tiny flecks of metallic film which changes color when viewed from different angles, see Figure~\ref{fig:200_500_ruble_ovi}. Colour-shifting inks reflect various wavelengths in white light differently, depending on the angle of incidence to the surface. OVI ink is extremely expensive and is generally used only in small areas. It needs to be printed in heavy weight and is sometimes printed using the silk screen process. Most common colour changes are brown to green (and viceversa) as well as red to purple. Ink feels almost embossed on the substrate which is due to the amount of ink required to get the required effect. Authors in~\cite{Chang2007} analyzes the reflected spectral signal of a banknote digits to extract its OVI properties. Extracting features for each of the ink colors at different angles, they expect counterfeit banknotes to not reproduce the same range of colors.

\emph{Holographic ink} is used for one of the most known overt features, which is the \emph{hologram} and its being used to protect a broad amount of documents like credit cards, see Figure~\ref{fig:hologram}. An Hologram incorporates an image with some illusion of 3-dimensional construction, or of apparent depth and special separation. They can be incorporated into tear bands in overwrap films, or as threads embedded into paper substrates. However, some hologram labels have been easily and expertly copied or simulated, and may often rely on hidden covert elements for authentication. Recently researchers have created a way to not only print chromatic holograms on any surface but also to create high-quality organic piezoelectric structures~\cite{keller2018inkjet,safaryan2018diphenylalanine}. Holograms are a type of OVD. OVD, based on diffractive optical structures, often without any 3D component is an image composed by an \emph{iridescent ink} which exhibits various optical effects such as movement or color changes. OVD can be created through a combination of printing and embossing. 
They are generally made up of a transparent film which serves as the image carrier, plus a reflective backing layer which is normally a very thin layer of aluminium. Other metals such as copper may be used to give a characteristic hue for specialist security applications. Extra security may be added by the process of partial de-metallization, whereby some of the reflective layer is chemically removed to give an intricate outline to the image, as can be seen on many banknotes. Alternatively the reflective layer can be so thin as to be transparent, resulting in a clear film with more of a ghost reflective image visible under certain angles of viewing and illumination. Partial removal of the metallic layer is a more restricted process and thereby increases both the level of security and the cost. 

Inside the semi-covert ink category, \emph{Thermochromic ink} is designed to be sensitive to temperature. It will appear or disappear at different temperature ranges. If you were to apply a thumb to a 15\degree C dark blue printed thermochromic spot the ink would disappear to nothing and as soon as you removed the heat source the ink would re-appear again. While it comes in a variety of temperature sensitivities, common temperatures available are 15\degree C, 31\degree C and 45\degree C. Before using this industrial ink, it’s vital to consider the temperature conditions to which it will be exposed from the time of imprinting through its lifecycle. In hotter climates might be needed higher temperature inks as the it could be invisible from the ambient temperature itself. Some are available as a permanent change. e.g. when it has reached a temperature the ink colour does not reverse.

Based on the amount of light receive, \emph{Metameric inks} corresponds to a pair of inks formulated to appear to be the same colour when viewed under specified conditions, normally daylight illumination, but which are a mismatch at other wavelengths. \emph{Photochromic ink} darkens, as the light level increases. It contain special chemicals which when exposed to ultra violet light e.g. sunlight, change from almost colourless to intense coloration. When removed from the source of ultra violet light, these inks revert to their colourless state. The photo-induced coloration commonly found in sunglasses may be used as anti-counterfeit markers on banknotes and documents such as passports.~\cite{higgins2003}. \emph{Phosphorescent inks} glow in the dark after having been exposed, for a variable period of time, to daylight. They are able to absorb the brightness and emit light even after the radiation, responsible for the fluorescence, is no longer present. The length of glowing time of the inks depends on the pigment type, the time of light exposed and the quantity of ink.

\emph{Reactive ink} is also referred as solvent sensitive. This type of ink can detect when there is an attempt to alter the document by a solvent or chemicals, such as bleach, alcohol or acetone. Reactive ink is usually found where variable data is printed on, these inks will run, change color, or cause a stain if an attempt to remove or alter information has been made. Reactive ink is commonly found in cheques, and is used on a printed watermark or fine guilloche artwork design. \emph{Fugitive ink}, also known as aqua fugitive inks, is designed to react similarly as reactive ink. Any form of alteration attempt (with water or an aqueous solution) on a security document with fugitive ink, the ink will run, causing it to smudge and become unreadable. Even wiping the finger with saliva on it across the printed background will make the printed pattern ink smudge. There exists combined \emph{solvent sensitive aqua fugitive inks} which combines both functions of reactive and aqua inks.

Lastly, in the covert ink category, many security printing inks rely upon the absorption of ultraviolet radiation and its re-emission as visible light. For that reason, to work correctly, some security printing ink designs needs to be printed on UV-dead or uncoated paper. If no UV brighteners are present in the substrate it will work at other documents~\cite{vuarnoz2008ink}. Additionally the printer can overprint varnishes and laminates, jointly with the security printing inks, to increase the difficulty to the counterfeiter. 

\emph{Invisible or fluorescence ink} security ink is the most known and common used, see Figures~\ref{fig:10_Ruble_invisible_ink},~\ref{fig:100_ruble_fluorescence_ink}. Designs printed with fluorescence ink, invisible in daylight or artificial light, becomes visible when exposed to a black lamp (or UV light source) and does not produce, as it is the case with phosphorescence, any after-glow  (persistence of fluorescence after switching off the Ultra Violet light). Invisible inks can be alcohol based or acetone based and either have white pigments or are without pigments. The composition of the ink can be controlled with additives to impact the response of the ink to a particular wavelength of ultra-violate or Infrared light. The ink must be applied to a UV dull substrate otherwise it will not be visible. As invisible ink is carried by the solvent it is relatively cheap and is available in many colours. UV inks are used in conjunction with a security background design to provide a higher level of document protection. \emph{Secondary Fluorescing ink} works in the same way as fluorescing ink however it will not glow or show under a black lamp unless an alteration or tampering on the material has occurred. For example, the secondary fluorescing ink will look green under UV light, but changes to red (secondary color) if an alteration has occurred. It is also possible to determine the fluorescence lifetime to discern the differences between genuine and counterfeit currency, using a two-photon microscope~\cite{chia2009detection}.  

At infrared spectrum it is also possible to search for local defect due deterioration  (like tears, stains or holes) along with the anti-counterfeit features~\cite{jin2008hierarchical}. Other works prefer to binarize and segment the image searching for IR patterns~\cite{chae2009study,lee2010image, bruna2013}.  

Predominantly used where numerical sequences or serial number security is important, \emph{Magnetic ink} contains minute magnetic flakes designed to communicate with electronic readers for document verification. Most common example of these is bank checks that use MICR (Magnetic Ink character Recognition) technology for highly sensitive data like check number, account number, and sort code of the bank. Magnetic ink is a type of \emph{Machine-readable inks} which at its name implies are the inks made to be only read by a determinate type of specialized equipment or solution. The main objective of these inks i to address the increased challenges of automated document security handling. Pulse eddy is an advanced electromagnetic inspection technology, that can be used to test magnetic inks or security threads~\cite{qian2011detection}. 

Among the various types of biometric personal identification systems, DNA provides the most reliable personal identification. \emph{Biometric inks} contain DNA taggants (uniquely encoded materials, like a fingerprint’s signature of identity) which are virtually impossible to duplicate and represent the ultimate marker for security purposes~\cite{HashiyadaDNAInk}. Special machines are designed that can read the tags, or the tags can be manufactured to react when a particular solvent comes in contact with them. Each batch of printed documents can contain different biometric properties. These are completely covert but require specialist methods to validate the authenticity. Table~\ref{tab:summary_inks_section2} presents a summary of the presented algorithms based on security inks.

\begin{table}
 \caption{ Summary of research works of anti-counterfeiting ink features for banknotes.}
  \centering
  \begin{tabular}{llll}
    \toprule
    Type                &   Feature            &    Method             & References \\
    \midrule
    OVI                 &   OVI digits   & spectral analysis   & \cite{Chang2007}       \\
    \midrule
    Fluorescence         &   fluorescence lifetime     & fluorescence spectra   & \cite{chia2009detection}       \\
    Magnetic ink         &   magnetic inks, Security threads    & pulse eddy   & \cite{qian2011detection}       \\
    \midrule
    UV                 &   paper substrate texture    & norm map                               & \cite{clarkson2009fingerprinting} \\
    \cline{2-4}
    \multirow{3}{*}{IR}                 &   paper substrate texture    & texture microstructure                         & \cite{voloshynovskiy2012towards,diephuis2013physical,diephuis2014framework,sharma2011paperspeckle, wong2017counterfeit} \\
             & Patterns    & binary segmentation & \cite{chae2009study,lee2010image, bruna2013} \\
             & Patterns and defects  & mesh of control points &  \cite{jin2008hierarchical} \\

    \bottomrule
  \end{tabular}
  \label{tab:summary_inks_section2}
\end{table}

\subsection{Security Printing}
\label{subsection:security_printing}

Offset Lithography and Intaglio press are two primary security document printing methods, differing between them by the solvents chemical composition used. Printing, the application of colour, consists of two main components: pigments, solvent. In addition modifiers and additives, like driers, waxes or anti-skinning agents are added to the ink to change its properties. Pigments are the ingredients that comprise the color of the ink. The ingredients are formulated from substances, found in nature or produced synthetically, which create the desired color when blended together in specific proportions. The solvent combines the pigment with the drying agent, responsible of speeding up the ink drying process and bind the pigment to the substrate. Wax additive improves the slip and scuff resistance of the ink, it also reduces the possibility of the ink to be transferred from one sheet to the back of another sheet. Anti-skinning agent adds the property of keeping the ink from drying too rapidly and skinning over in the ink fountains of the printing press. The application and varying composition of pigments and solvents during the printing process of a security document is indeed a security feature. 


Although not a secure printing process in its own right, \emph{Offset Lithographic printing} is the most common commercial printing process used in secure documents. Often used as a security feature, background printing represents a great portion of the overprinting of any security document. The image on the printing plates is defined by raised areas. The raised areas are inked up and instead of being transferred straight to the paper the design is transferred or offset to a rubber blanket. The print is then transferred from the blanket to the paper. Lithography is used for most of the background printing and is the first to be printed. Lithography requires a different printing plates for each color of components present at the background design. The ink is oil based and the resulting print is flat crisp, sharper line on edges, and a brighter overall results than Intaglio press printing.

\emph{Intaglio printing} plates have the image area etched into them. The plate is inked up and wiped, having only ink in the recessed areas. The ink from the plate is then directly transferred onto the paper substrate at a high temperature and pressure (with one metric ton per cm linear). The printing conditions are such that the paper is sucked into the recesses in the plates and deformed. The ink then sits on top of the paper deformations and hence a tactile effect results from a combination of paper deformation and ink thickness. Ink dries by evaporation, being the drying longer than Offset Lithography. This longer time of drying causes an ink feathering effect, where the edges appear to run. This printing technique and the capillary effect lead to embossing of the substrate and the tactile high amount of ink. Intaglio printers can produce non-uniform quality standards and regional production disparities of an Intaglio print appearance. This difference in the look of an intaglio zone, may confuse security document reviewers and help counterfeiters to produce an similar intaglio respect the genuine one. To reduce this variability and quality deviations, Intaglio quality analysis and measurement can be performed\cite{hofmann2014new,funkb2016intaglio}. To describe quality levels they use Intaglio line discontinuities, bleeding and inner holes in the lines as well as large areas without ink.

\begin{figure}
     \centering
     \begin{subfigure}[b]{0.3\textwidth}
         \centering
         \includegraphics[width=\textwidth, height=3cm]{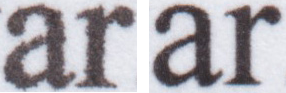}
         \caption{Inkjet-printed vs Laser printed}
         \label{fig:inkjet_vs_laser}
     \end{subfigure}
     \hfill
     \begin{subfigure}[b]{0.3\textwidth}
         \centering
         \includegraphics[width=\textwidth, height=3cm]{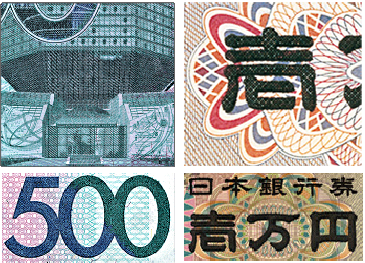}
         \caption{Intaglio printing}
         \label{fig:intaglio-printing}
     \end{subfigure}
     \hfill
     \begin{subfigure}[b]{0.3\textwidth}
         \centering
         \includegraphics[width=\textwidth, height=3cm]{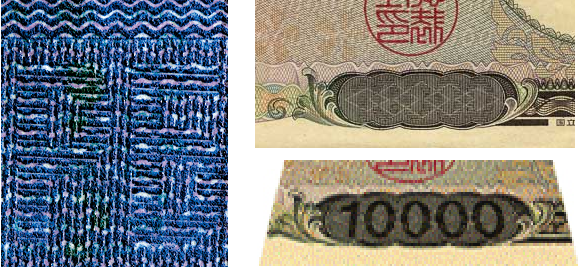}
         \caption{Latent images}
         \label{fig:latent}
     \end{subfigure}
        \caption{Examples of anti-counterfeit security printing features. (a) Difference between inkjet and laser printing. Tails or satellites are formed in the inkjet because the diffusion speed of the ink drops is quicker than that of the fused toner on the paper. Laser printing has sharper countours. (b) Intaglio zones in Rubles and Yens. Image designs of the values, portraits or main images on banknotes are usually thickly printed to give a notable texture. (c) Latent images. When tilting, the denomination of the Japanese banknote appears on the right image. The left image corresponds to a zoomed latent region in the Ruble banknote. "RB" in Cyrillic letters are visible when tilted and held up to the light.}
        \label{fig:security_features_based_printing}
\end{figure}

Intaglio printing has a distinctive feel to it and can also be checked easily, simply by running a finger over the printed page, see Figure~\ref{fig:intaglio-printing}. Intaglio printing process can produce \emph{latent images} as a counterfeit security measure. Latent images can be viewed when tilted and illuminated using side/oblique lighting. When viewed straight on, a latent image reveals nothing but lines, see Figure~\ref{fig:latent}. They are composed by patterns of raised lines at right angles. The fine raised ink pattern is rendered variable in contrast to the foreground. Counterfeits made using the intaglio process had been seen on rare occasions due intaglio presses are far more expensive than ordinary offset, typographic or lithographic presses, which yield inferior counterfeits. Moreover the tactile effects in particular are hard to reproduce. 
Different works compares the characteristics of motifs printed with Intaglio and other printing processes, such as offset and use these differences for counterfeit detection~\cite{Glock2009,roy2015machine,Lohweg2012,Lohweg2013}. 

Other commercial printing processes such as letter-press, flexography, gravure, and screen printing are not specific to secure documents, however they are often used for document numbering, laminate printing, and security feature inclusion.


\emph{Personalized printing} corresponds to the printing of variable information between security documents of the same type, thereby allowing individualize the document. Banknotes are classified by their value, series and country. Usually, the only intra-variance between the same group of banknotes corresponds to the \emph{serial number}. Serial numbers  provides security and identity to the notes on which they are printed. In the case of identity documents all personal information requires personalized printing. High-volumes of unique information printing is required, hence printing methods that are readily available, versatile, and cost effective are mandatory. These requisites are accomplished by \emph{desktop production} or commonly named as \emph{desktop printing} and are available and affordable to the ordinary person. Depending on the document substrate type, various desktop printing processes may be employed. 

Most documents are produced with inkjet, laser, thermal transfer, or dye diffusion thermal transfer. Laser engraving is used for synthetic documents. However is it possible to find traditional processes as typewriter and dot matrix machines at older issued IDs, which does not have expiry date or are still valid nowadays.  
Inkjet and toner printing processes, together with typewriters and dot matrix printers, are restricted to print directly on paper based documents, see Figure~\ref{fig:inkjet_vs_laser}. An exception to this direct transfer of the ink to the substrate is its printed onto the reverse of a laminate adhered to the substrate. If a photo needs to be added onto an ID, a physical patch should be added to the traditional printing processes, meanwhile with inkjet and toner it is possible to print text and images directly at the document.  
Referring polymer substrate printing, meanwhile dye diffusion thermal transfer combination only prints correctly on polymer substrates, more general thermal transfer printing can be used on both polymer and paper documents. Both of this processes, likewise inkjet and toner, allow to print either black and white or color photos directly onto the substrate. 

A well-known technique to print personalization information is \emph{laser engraving}. This technology does not involve the use of inks and can only be used on "laserable" materials like synthetic polymer documents. The carbon sensitized layer within the polymer substrate, into which data is engraved, is made with plastic containing particles that are laser sensitive. A laser beam burns the particles of this layer, printing the personalized information. As a consequence of the carbon particles, printing can only create black and white images. This laser-based process creates flat and/or raining printing, into the thickness of the document, impossible to remove and adding further security to the document. In~\cite{dewaele2016forensic} variations of statistics along edges between a printing press, a laser and an inkjet can be differentiated to identify counterfeit bills using an ordinary mobile device.  

\emph{Multiple laser image} (MLI) based on the laser engraving technology is widely used as a security feature on ID documents. MLI as a single/separate element and in a simple format is regarded as having lost its strength and security role. However, the combination of MLI with offset printing, overt and covert security features and utilizing the latest advances in lamination plate technologies, has transform MLI in one strong authenticator~\cite{li2015optical,gemalto-passport-security-design}. 

\begin{figure}
     \centering
     \begin{subfigure}[b]{0.3\textwidth}
         \centering
         \includegraphics[width=\textwidth, height=3cm]{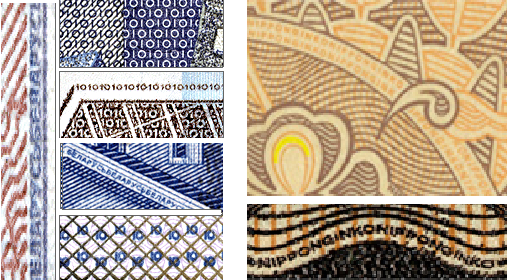}
         \caption{Microprinting}
         \label{fig:microprinting}
     \end{subfigure}
     \hfill
     \begin{subfigure}[b]{0.3\textwidth}
         \centering
         \includegraphics[width=\textwidth, height=3cm]{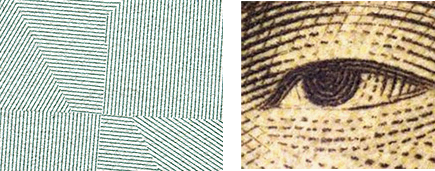}
         \caption{Ultrafine-line printing}
         \label{fig:ultra-fine-line-printing}
     \end{subfigure}
     \hfill
     \begin{subfigure}[b]{0.3\textwidth}
         \centering
         \includegraphics[width=\textwidth, height=3cm]{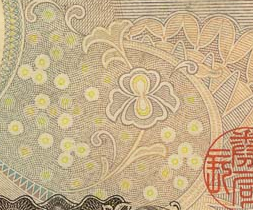}
         \caption{EURion constellation}
         \label{fig:euro-constellation}
     \end{subfigure}
    \caption{Examples of anti-counterfeit security printing features. (a) Microprinting. Is possible to read text when the region is magnified using a tool with magnifying glass. Ruble image regions (left) has written "NBRB" and "Belarus" in Cyrillian text. Yen images (right) regions contain written "NIPPON GINKO" meaning "Bank of Japan". These small characters can hardly be reproduced without the original manufacturer printer. (b) Ultrafine-line printing. Such fine lines act as anti-replica zones for color copy machines or by ordinary printing equipment. (c) EURion constellation. These screen traps helps imaging software to detect this yellow dots and block the user from reproducing security documents. On Japanese yen, these circles sometimes appear as flowers.}
    \label{fig:security_features_based_printing_2}
\end{figure}

Nowadays all major commercial manufacturers of color laser printers have entered a secret agreement with governments to ensure that the output of those printers is traceable~\cite{EFF_tracking_yellow_dots}. The U.S. Secret Service admitted that the tracking information objective of this measure to identify counterfeiters. Each printer may contain some kind of forensic tracking code, visible or not to the naked eye. The most famous subtle pattern are the yellow dots covering the document. Those dots are a microscopic code that allows government agencies to trace the documents back to the printer which create it. Researchers lead by the Electronic Frontier Foundation (EFF) cracked the code of dots on documents made by Xerox, company who pioneered this technology since 1984, to assuage fears that their color copiers could easily be used to counterfeit bills~\cite{EFF_cracking_yellow_dots}. These anti-counterfeit marks are known as \emph{screen traps}, causing most of the scanners, copiers, and desktop publishing software, to fail to reproduce the document when detected. Found in most of the banknotes, an example of this yellow dots, is the \emph{EURion constellation} found on Euro banknotes, consisting of a pattern of five small circles, in exact distance apart, size, proportion and colour, which is repeated across areas of the banknote at different orientations, see Figure~\ref{fig:euro-constellation}. The mere presence of five of these circles on a page is sufficient for some colour photocopiers to refuse processing. Constellation should be of exact measurements and colour properties, if the constellation is altered by even a small variation, copying is possible~\cite{verikas2011advances}. Similarly to yellow dots, authors in~\cite{kim2007data} present a method for data hiding in any bicolour printed documents, which can contain characters, drawings, schematics, diagrams, cartoons, but not halftones. They pseudo-randomly distributing tiny black dots to embed the message. When the security trap is produced with thermal ink jet, dry electrophotographic, and liquid electrophotographic digital printers, it is possible to calculate its security strength~\cite{sturgill2008security,simske2009new}. Repeated line patterns, two-dimensional (2D) bar code reading, and authentication of a color deterrent (color tile), prove to be effective to measure colour fidelity. The optimum selection of printing strategy, print technology, substrate and printed pattern may reduce the options of a counterfeiter. 

\emph{Serial numbers} serves to identify uniquely each banknote printed and in circulation. It is a simple, cheap and effective security measure designed to make life difficult for the counterfeiter. In IDs documents, identity numbers are the synonyms for the serial number and can vary for each country. Additionally a format and limit ranges are added to increase the level of security. These numbers are stored in databases, accessible through ATMs and government systems, which can quickly notice if there are duplicates simultaneously at different geographically locations. Several works include serial number authentication in their pipelines~\cite{wenhong2010application,zhao2010study,feng2013extraction,feng2014automatic}. Additionally the age of ink of the serial numbers can be determined and consequently compared with the date it was issued. A essential zone for some IDs, like passports, is the Machine Readable Zone (MRZ). Similarly to serial number, MRZ contains identity number or/and passport document number, which also follows some rules and limit ranges. Those can be stored in a database for authentication. Furthermore, it also contains cheksum digits along with other bio-data information which can be compared with the printed data at other zones of the ID for detecting counterfeits~\cite{kwon2007recognition}. 

\emph{Microprinting} is another personalized anti-counterfeiting technique used in both banknotes and ID which is difficult to reproduce by digital methods. The patterns or characters are printed at a scale that is only visible through magnification with a magnifying glass, or microscope. To the naked eye, microprinting may appear as a solid line, see Figure~\ref{fig:microprinting}. Counterfeiters who tries to replicate these patterns using a photocopy or image scanning usually translates as a dotted, solid line or very low quality text on the counterfeit item when it is printed. Usually microprinting personalizes personal information like dates, words, abbreviations or serial numbers.  

The \emph{Typography} used at each of the fields of banknotes and IDs are also a security feature. The font, style and size of characters is a very distinctive feature in security documents~\cite{bertrand2015conditional}. Most of the printable fonts in IDs use proprietary, custom-designed typefaces. The fact that these typefaces might look like some commercial or publicly available typefaces is coincidental. Intrinsic characters and fields features, which we name \emph{Intrinsic}, for the bio-data printed characters like the layout, alignment, skew and shapes, among others are also used as security measures to catch inexperienced counterfeiters~\cite{bertrand2013system,satkhozhina2013optical,van2013text}.


\emph{Complex and ultrafine-line} printing designs drawn with extremely fine lines are printed on the security documents to prevent the rendering of these intricate patterns, see Figure~\ref{fig:ultra-fine-line-printing}. Even if high quality desktop equipment may be capable of render them, when magnified the area of the patterns the edges begin to blur. The intricate level of detail of the complex geometrical designs achieved by Intaglio and Offset Lithography remains as a secure anti-counterfeit measure against counterfeiters.

An example of geometrical complex printing designs are the \emph{Guilloches}. Guilloches are patterns of subtle thin lines interwoven according to the rules of geometry. The wavy decorative lines and graphic patterns which composes them are primarily used on banknotes and ID documents. It works using a mathematical shape known as a hypotrochoid, which is the equation for the fixed point of a circle rolling around inside a larger circle (same concept as a spirograph works). 

\emph{Moiré patterns} are designed to interfere with halftone screens used in the printing industry and applied for digital photocopying or digital scanning prevention of security documents. Moiré patterns are a type of screen trap, similar to the yellow dots. This type of screen trap produces strong artifacts when a document is counterfeited using any standard halftone-based colour reproduction system such as offset printing. A new visible observable pattern is created when the individual structures are superimposed, being the new pattern different from the original structures. Moiré effect is an optical phenomenon generated by the interference between two different superimposed periodic structures, like line gratings or dot screens. The new observed image is extremely sensitive to small variations in the original layers, hence can be used as a anti-counterfeit feature. Three general different types of moiré patterns are commonly found named 2D, 1D and pseudo-random, based on their moiré intensity.

\emph{Vignettes} are decorative and intricate designs which resembles to pieces of art, used to to increase the security of the documents. Nowadays is not a high security feature due the advances of scanning and printing technology. Banknotes have started to combine vignettes with other security features, like Australian polymer banknotes, which embeds a vignette inside of a clear window, creating a robust anti-counterfeit feature. 

Summarizing security printing, see Table~\ref{tab:tab:summary_printing_section2}, some works uses the whole security document to find differences at different regions after applying a single feature extraction. These regions usually correspond with a security feature, like the vignette, microtext, watermark or security thread, to cite some. The feature extraction can be just binarize the whole document and then try to find discrepancies respect a reference genuine document~\cite{prasanthi2015indian,alshayeji2015detection}. The authors in ~\cite{berenguel2016banknote,aberenguel:evaluation, aberenguel:econterfeit} also use different regions of the IDs and banknotes, to evaluate the background texture. These regions are placed in the zones containing Guilloches, Vignettes and Intaglio. 

\begin{table}[h]
 \caption{ Summary of security printing techniques studies of anti-counterfeiting features for banknotes.}
  \centering
  \begin{tabular}{llll}
    \toprule
    Type        &    Method            & References \\
    \midrule
    Offset Litographic  &   edge variations, respect inkjet and laser   & \cite{dewaele2016forensic} \\
    Intaglio design     &    texture quality   & \cite{Glock2009,Lohweg2012,Lohweg2013} \\
    Serial Number       &   binarization       & \cite{wenhong2010application,zhao2010study,feng2013extraction} \\
    MRZ       &   binarization, template matching       & \cite{kwon2007recognition} \\
    Typography & conditional random field model                     & \cite{bertrand2015conditional,satkhozhina2013optical} \\
    Typography, Intrinsic & hu moments, character size/alignment/axis inertia                     & \cite{bertrand2013system} \\
    Guilloche, Intaglio, Vignettes & combination texture features               & \cite{berenguel2016banknote,aberenguel:evaluation,aberenguel:econterfeit} \\
    Layout & text-line skew and alignment                     & \cite{van2013text} \\
    Microtext, Background patterns      &  binary segmentation                      & \cite{prasanthi2015indian} \\
    Background patterns           &  slicing, edge detection, binary segmentation                      & \cite{alshayeji2015detection} \\
    \bottomrule
  \end{tabular}
  \label{tab:tab:summary_printing_section2}
\end{table}

This whole section was based on anti-counterfeit security measures to detect counterfeit documents. Anti-counterfeit measures are essential to reduce the cost fraudsters of security documents are causing to the society, see section~\ref{section:effects_society_and_document_experts}. Authorities alert the public to remain vigilant for people attempting to pass off counterfeit banknotes, and to call police should they be presented with what they suspect may be fake currency. However, human inspection of this bogus banknotes is prone to errors. Also ID theft industry protection have boomed over the last years, offering services that can authenticate an ID received from a customer, who has digitally scanned the document. This companies has to implement algorithms to automatically check the sent document is a photocopy or not to provide a minimum level of security. In addition to photocopying and scanning, in next section we detail more types of attacks and vulnerabilities in security documents. We also focus on approaches to protect digital documents from tampering.

\section{Types of attacks and vulnerabilities}
\label{section:types_of_attacks_vulnerabilities}

Duplication, imitation and mutilation are the most common types of attacks to forge a document or banknote, see Table~\ref{tab:types_of_attacks}. The three types of attacks are detailed next. 

\begin{table}
 \caption{Types of attack in security documents.}
  \centering
  \begin{tabular}{lll}
    \toprule
    Mutilation          &   Imitation                           & Duplication                   \\
    \midrule
    Ink removal         &   Desktop production                  & Photographing                 \\
    Precision cutting   &   Commercial production               & Photocopying                  \\
                        &   Intaglio/Lithographic production   & Scanning                      \\
    \bottomrule
  \end{tabular}
  \label{tab:types_of_attacks}
\end{table}

\begin{itemize}
    \item Mutilation. The most aggressive technique of modifying a genuine document or banknote, in which the original parts are removed or replaced. Usually banks exchange full value for mutilated banknotes which a portion is missing or which is composed of more than two pieces. This fact is used for the counterfeiters to disguise other types of forgeries in the banknotes. In documents is a rare attack because the official identity documents are more prone to be replaced when damaged.
         \begin{itemize}
            \item \emph{Ink removal}. Removing ink can be use to modify the value of a lower value banknote to a higher value preserving the original substrate of the banknotes or alter the information at some documents. From light amounts of chemicals such as bleach, Acetate or Acetone present in most nail polish removers, chlorofluorocarbon existing in hairspray and denatured alcohol or witch hazel present in after shave lotion to cite some is helpful to remove marks in any surface. Banknotes can also be ink-stained agains a stole attack. When criminals open a protected cash container, an ATM or a safe in a cash transportation vehicle, an anti-theft device known as intelligent banknote neutralization system (IBNS) can be activated and stain the whole banknote or some parts of it make in it unusable. When a banknote is stained by an IBNS, the security ink soaks into the banknote flowing from the edges to the center leaving a characteristic pattern. When the criminals tries to use chemicals to remove this security ink the original colours could be also altered, and some security features may be damaged, or may even disappear.
            \item \emph{Precision cutting}. ATMs can also protects itself against an attack, using glue to fuse all the banknotes into a solid brick and render the cash unusable. If the thieves try to peel off individual banknotes, they tear into pieces. Banknotes or documents can be cut into precise vertical strips and then joining with clear adhesive or glue to produce forged new ones. In documents the strips can be used to alter some biodata information. Although difficult to detect, under close inspection, image edges tend to slightly deviate from the original image. 
         \end{itemize}

    \item Imitation. This term correspond to the ability to fabricate new documents or banknotes with the available technology to common consumers. The equipment that can be obtained without the supervision of the government or special authorities.
        \begin{itemize}
            \item \emph{Desktop production}. Common office printers can produce resolutions of $300$ dots per inch (dpi) or more. Nowadays the casual or low funded counterfeiter, can acquire the necessary equipment within a reasonable price which can already reproduce high quality documents and banknotes passable at the first glance. Most photo IDs are printed by digital thermal transfer, color is transferred from a single-use ribbon to various kinds of receptor materials by this process. Intaglio and Offset Lithographic methods can not be matched and under microscopic inspection, microtext and Guilloché patterns are not rendered with the required quality. 
            \item \emph{Commercial production}. Digital printing business has last-generation commercial equipment at their disposal capable of reproduce microtext and Guilloché patterns with higher precision. Image fidelity and image quality are superior than desktop production due it uses better inks and higher resolutions. The substrate of the document or banknote imitates closer to the genuine one in weight, thickness, fiber texture and surface. Security components such as watermarks, security fibers and security threads within the substrate are not replicated because this knowledge is kept secret by the issuing authorities. 
            \item \emph{Intaglio/Lithographic Production}. Usually both Intaglio and Lithography are printmaking processes only known by the legitimate issuing authorities, however criminal organizations with enough resources have available the necessary equipment and knowledge of the inks and printing design to reproduce the original document. Spotting the counterfeits is extremely difficult in these cases. Forensic document examiners can rely at observed uncharacteristic anomalies at known legitimate printers. Moreover, security fibres, threads and watermarks can provide some clue of forgery if they are misaligned or deviate from the original document. 
        \end{itemize}
        
    \item Duplication. The idea is to duplication is to fabricate an identical copy of the original security document in all of its aspects. Duplicate means to make a perfect copy and absolutely identical to the original document, which would be the ideal case for the counterfeiter. However, finding replicas of security documents is much more common. A replica means to get a copy that is almost the same as the original, but not quite the same, it is always slightly different from the genuine at least in terms of its identity.
        \begin{itemize}
            \item \emph{Photographing}. Usually the forger starts a new counterfeit inside a dark room, where acquires the legitimate document with a high quality photograph. A separate photograph is taken of each shade used on the document, as well as the pattern on the back. Each photo is stored in film or photographic negative, where the pattern of the document is transparent. A machine prints the pattern on a thin plate using light. The light going through the transparent bits prints the pattern onto the plate. For each shade a separate plate is created. The negative fails to hold all the detail of the original, so it has to be touched up by hand. Finally each plate is inked and printed on paper or other material. The produced counterfeit has vivid colors which at first glance are identical to the genuine document. However by touching the substrate is it possible to appreciate different tactile qualities.
            \item \emph{Scanning}. For an affordable price it is possible to buy scanners able to scan a document with resolution over $2400$ dpi. These scanners are often referred to as copy-dot scanners because they attempt to copy all of the halftone dots in the original. The purpose of greater scanning resolution is to modify the images with editing software and achieve a greater precision of the modification. Spotting the tampered document will depend on the printing operation and the quality of the ink and substrate.
            \item \emph{Photocopying}. This type of forgery is only performed by the casual counterfeiter and it is the easiest to spot. Usually the ink and paper quality correspond to standard office supplies, which makes the reproduction differs greatly from the genuine and easily to differentiate by the substrate and dull colors. Along section~\ref{section:anti-counterfeit-measures}, we have introduced different anti-counterfeit security features that can be used to authenticate a security document. A color copier or scanner can copy a document only at one fixed angle relative to the document surface. In a banknote inspection is possible to search for most of this features as a mean to determine if a note is genuine. Most of the security features will not be present in a photocopy. On the other hand, when validating an ID document using a single camera-base acquisition that has been sent online, most of the security measures can not be inspected. Some elements of minor importance, in terms of security feature level, are more or less visible on photocopies, and can be used for authentication such as stamps, holographic films, perforate numbering, the paper embossing, the typography. However the reliability of such methods will depend on the quality and resolution of the photocopy. The rest of the security features can not be controlled or detected from a photocopy. The same applies behaviour applies to the scanned images, however if the resolution of the scanned image is sufficient, other security measures can be checked, as the motifs, perforated image, Intaglio or background complex printing or texture designs.

        \end{itemize}
\end{itemize}

\subsection{Digital Tampering}
\label{section:tampering}

The previous section~\ref{section:anti-counterfeit-measures} corresponds to a summary of the most common used anti-counterfeiting techniques against the tampering of security documents. Typically fraudsters try to create a replica or duplication of security documents in a physical material format, to ultimately use these counterfeit for criminal deception purposes. Having the existing physical document is mandatory to show in person to the corresponding authority or seller in order to buy goods or contract services. Although the last statement is compulsory for banknotes, it does not always applies for identity authentication and verification. Nowadays most of the services or products required for the general population, such as opening a bank account, applying for a loan, renting a car, checking-in in a hotel, etc. are easily available to contract through Internet. The companies who offers this services, needs a genuine online identification of the interested client before formalizing the contract or provide the service. The process of on-boarding a new client needs to be fast and seamless to obtain as many clients as possible in the shorter amount of time. Generally the client is asked to acquire his identity document with a smartphone or digital camera and send it online. In this case if the client is an imposter who has created a tampered document by most of the attacks explained previously in this section, most of the anti-counterfeiting techniques from are rendered useless once the document is transformed digitally to a single image. 

As technologies used to digitally authenticate people over the decades have advanced, so too have the techniques attackers find to trick or bypass digital authentication. Fraudsters may modify some parts of its own personal data or impersonate completely other citizen, previously stealing his ID to afterwards replace some parts such as the photo. Typically forgers find is an easier option to replace small portions of real existing personal data information printed on a document than preparing a fake ID document from scratch. Most common forgeries replace the photo or change the number or letters containing the bio-data information, like changing the expiry date in a residence permit card. Even common users can produce high skilled documents forgeries, due to the availability of low-cost, high-performance computers, and the emergence of powerful software for processing and editing images. It has become relatively easy to manipulate or edit digital images even for non-professional users. \emph{Digital Tampering} definition is the procedure of replacing the content within a region of the original image by some new content using editing software. 


\begin{table}[h]
 \caption{Common tampering types categorized according the manipulation operations and image source.}
  \centering
  \begin{tabular}{llccccc}
    \toprule
    Type & Actions & Single image source & Region duplication & Tampering Objective \\
    \midrule
    \multicolumn{2}{c}{Copy-move} & \cmark  & \cmark &  Object removal \\
    Splicing      & Cut-paste     & \xmark & \cmark &  Object addition \\
    Inpainting    & Erase-fill    & \cmark & \xmark &  Object removal \\
    \bottomrule
  \end{tabular}
  \label{tab:tampering_types}
\end{table}

The three most common tampering types are: \emph{splicing}, \emph{copy-move} and \emph{inpainting}, see Table~\ref{tab:tampering_types}. \emph{Splicing} is a technique of creating an image by combining two different images. In image splicing a majority part of one image is used. The objective is to achieve the impression the new foreground object is part of the background. In \emph{Copy-Move} type of forgery, a part of the image is copied and pasted onto another part of the same image to hide some object or some detail. These type of forgery is used to hide so information or alter the bio-data with copying the letters and numbers from other text fields with the same typography. When the part duplicated, using a single image, is removed from the original location and filled, this technique is called \emph{Inpainting}. The filling operation is usually performed with gradient techniques to achieve realistic backgrounds. Recent image editing software suggests, for the three tampering types, to use neighboring patches or pixels within the original image to replace the target region because using these patches is easy and more likely to achieve smooth filling effect than using patches from another arbitrary image.

\subsection{Passive tampering approaches}
\label{subsection:tampering_detection_approaches}

Different surveys has been published on image forgery detection~\cite{zampoglou2017large,zheng2019survey,teerakanok2019copy}. The works cited in those surveys are mainly evaluated against very large collection of forgeries datasets collected from various Web and social media sources. Currently there is no publicly available tampered identification document dataset, read section~\ref{section:datasets} to understand the reasons it does not exist. However the techniques applied at some of these works can be transferred to the digital identity tampering discovery. 

Identity documents are going to be acquired in a open-world scenario. \emph{Passive tampering} techniques aims at verifying the authenticity of digital images without any a prior knowledge, like the acquisition device or identity document layout. Passive detection algorithms exploit the artifacts and inconsistencies to distinguish between pristine and forged areas in the image~\cite{wang2009survey}. Among these algorithms statistical methods, based on pixel-level analyses, are the most common. These statistical methods can follow a model-based or a data-driven approach. Model-based methods use features like lens aberration~\cite{yerushalmy2011digital}, color filter array (CFA)~\cite{ferrara2012image}, JPEG artifacts~\cite{pasquini2017statistical} or camera response function~\cite{chen2017image} to build a mathematical model to detect the tampered areas. 

Data-driven algorithms are evaluated on the noise residuals of the image. Noise residuals can be obtained applying high-pass filters in the spatial or transformed domain, as Fourier~\cite{verdoliva2014feature}, DCT~\cite{he2012digital} or Wavelet~\cite{farid2003higher}. Authentic scanned text documents may contain multiple, similar-looking glyphs (letters, numbers, and punctuation marks). In~\cite{abramova2016detecting}, the authors study the impact of copy-move existing algorithms to this scenario, showing that under specific threshold and parameters values, the block-based methods have a modest performance. They also propose an analysis framework for detecting copy-move tampering in text images, joining OCR character characteristics like weight, size, style and roughness with the copy-move algorithms focused on the background. 

When a uniform source light falls on a camera sensor, each pixel should output exactly the same value. Small variations in cell size and substrate material result in slightly different output values. Photo response non uniformity noise (PRNU) stands for the differences between the true response from a sensor and a uniform response~\cite{lukavs2006detecting}. PRNU is caused by the physical properties of the sensor, it is almost impossible to remove entirely and is usually considered to be a normal characteristic of the sensor. Different works demonstrate PRNU patterns are a good option for identifying and localizing forgeries~\cite{chierchia2014bayesian,chakraborty2017prnu}. A drawback of this approach is that PRNU patterns must be estimated for each camera model, which uses a specific sensor type, requiring a large number of frames from the target camera model to obtain realiable results. 

Authors in~\cite{qian2015deep} use steganalysis to suppress the scene content and force the network to work with noise residuals using a deep learning model. In~\cite{zhou2018learning} a two-stream model network, where the first network learns the noise residuals and the second is a general purpose network. Authors in~\cite{bappy2017exploiting} propose a localization framework using an hybrid CNN-LSTM model to learn the boundary discrepancy between pristine and forged regions. In~\cite{bondi2017tampering} identifies different camera models using a CNN that compares image patches, however it requires the camera models to be in the training set. A Siamese network can learn the EXIF metadata, to create model that distinguishes patches from different camera sources~\cite{koch2015siamese,huh2018fighting}. Following this idea, Noiseprint exploits image content and camera model information~\cite{cozzolino2019noiseprint}. It can detect most of the tampering types. Noiseprint has been successfully used in forgery localization under a supervised setting and in video forensics~\cite{cozzolino2018camera,cozzolino2019extracting}. ManTra-Net is a novel that exploits a long short-term memory model to asses local anomalies~\cite{wu2019mantra}. 

\subsection{Digital watermarking}
\label{subsection:tampering_watermarking}

The previous section establishes a principal constraint which is the unavailability of the original image. When IDs are manufactured it is possible to acquire a digital image of the original genuine document. If \emph{digital watermarking} is applied to this image, it can be used for secure digital transactions. The digital watermaked image can also be reprinted embedding the code physically in the document. Digital watermarking are the techniques that hide information, for example a number or text, in digital media, such as images. The content of the digital data is manipulated to embed the hidden secret digital information, called "watermark". The pixel values modified and quality degradation in the watermarked image must be unnoticeable by human eye. Moreover, the watermark should be robust to resist manipulations or possible attacks on the digital image, such as lossy compression, scaling, cropping, among others\cite{kumar2018recent}. Furthermore, the hidden information should be possible to be detected or recovered, with the objective to verify the authenticity of the digital image. 

At early stages of computer vision to prevent counterfeit digital tampering attempts, the most notorious works were based in Discrete Wavelet Transform (DWT) and Singular Value Decomposition (SVD), \cite{cox2002digital,liu2002svd}. A watermark is embedded in host images at these approaches in spatial or frequency domain. 

Authors in~\cite{perry2000digital} inserts a unique personal code inside the photo to figure out fraud photocopy ID documents. The forgery is detected when the embedded data is compared with the ID information on the document. Also in in~\cite{thongkor2012digital} a ID personal number watermark is embedded into owner picture, the difference is that they print the photo watermarked on the ID. Later, with a camera-based acquisition of the printed ID, the watermark can be extracted and checked if belong the ID personal number corresponds with the one at the document. Projective transformation registration technique is applied to minimize perspective, rotation, scaling and translation distortions. 

Printing a digital watermark to a physical ID card and its posterior scanning for authentication may introduce random noises into the images, named printing and scanning distortion (PS). Distortion can appear at pixel values and the geometric boundaries of the scanned image. In~\cite{ibrahim2010adaptive} adaptative watermarking using matrix of regulation factors is proposed to remove the PS noise in the watermark before extract it. Another possibility is to combine different hiding technologies like digital watermarking, 2-D bar codes, copy detection patterns and biometric information to protect ID documents against several types of forgeries~\cite{picard2004towards}. Recent surveys in digital watermarking can be found in\cite{shih2017digital,singh2013survey,singh2016digital}. The later presents the use of anti-forensics sections, where they explain how the forgers hide the tampering as a result of the fingerprints study that might be introduced due to their use.

In most cases digital watermarking techniques can only be applied by the manufacturers of the original ID document. Furthermore, digital watermarking is weak against three types of attacks: removal, cryptography and protocol attacks. The removal attacks try to remove all watermarking in the document. The Cryptography attacks aims to alter the watermarking and the protocol attacks objective is to attack the watermarking applications.

\section{Approaches, methodologies, and techniques}
\label{section:approaches-methodologies}

In this section we summarize most of the algorithms presented in section~\ref{section:anti-counterfeit-measures}. We have divided in three steps the presentation of these algorithms: preprocessing, feature extraction and classification. These are the usual steps followed to authenticate a banknote.

\subsection{Preprocessing}
\label{subsection:approaches_preprocessing}

Most of the presented works for counterfeit detection need as an initial stage to have the banknote correctly cropped. Removing the background from the object that needs to be inspected, will contribute to better accuracy results of authentication. A simple cropping of the document can use Hough Transform on a Canny edge detected image. Afterwards registration can be done with Template matching of relevant patterns at the document. These are simple algorithms, which can be certainly improved with better approaches. However the registration of the document is outside of the scope of this work and we presuppose the cropping is already done.

Preprocessing step is commonly used as a prior to descriptor calculation in order to normalize illumination differences from the acquisition devices such as scanners, cameras, smartphones, etc. It is important to improve the quality of the sourced image and at the same time do not remove important information printing information which could lead to the counterfeit identification~\cite{mandridaketowards}. Banknotes and IDs surface can also be soiled by dirt and sebum from users hands. Banknotes are more easily contaminated due to their wide circulation. The environmental acquisition conditions can also introduce variations in the aspect of the acquired image, such as the exposure, brightness, contrast, etc. To address this, noise removal prepossessing techniques are introduced as a first step in the processing pipeline. Histogram equalization or storing a brightness map to normalize with the test images, is used for brightness normalization and contrast enhancement~\cite{yeh2011employing, Bhavani2014,bruna2013,berenguel2016banknote}. Authors in~\cite{mandridaketowards} presents a preliminary study on the difference between using gradient filter for illumination normalization for IDs background analysis. Author in \cite{elkasrawi2014printer} are interested in the noise produced by the laser and inkjet printing techniques. They filter the printed area calculating the Otsu’s threshold and getting the median gray-level for the background pixels from the original image.

Most of the works directly works with the RGB color space from the input image. If a cited work no color space is specified RGB is what they use. Other authors, to reduce the input dimension authors may adopt other colorspaces and only use one component. Only the Y component from the YIQ color space is used in \cite{yeh2011employing, Bhavani2014}. The b* component from L*a*b color space is used for analyze microletter in~\cite{roy2015machine}. Single channels of RGB, HSI and L*a*b colour spaces are compared, for complexity reduction in~\cite{Chambers2015}. They achieve higher accuracy if they use an average of RGB channels for their dataset. 

Authors may partition the image into different patches, scattered at the image input. The objective is to reduce the dimension of the input image, preserving global information~\cite{yeh2011employing, Bhavani2014, Desai2014ImplementationOM, berenguel2016banknote, aberenguel:evaluation, aberenguel:econterfeit}. The features are then extracted at each patch and combined in a later stage. In~\cite{alshayeji2015detection} original grayscale images are decomposed into their equivalent 8 binary images, claiming is useful in analyzing the relative importance contributed by each bit of the original image. Table~\ref{tab:preprocessing} summarizes
the preprocessing algorithms presented in this section.

\begin{table}
 \caption{Preprocessing techniques in the counterfeit flow.}
  \begin{center}
  \begin{tabular}{cccc}
    \toprule
    Task  & Method & References \\
    \midrule
    \makecell{Brightness normalization and \\ contrast enhancement}   & Histogram Equalization & \cite{yeh2011employing, Bhavani2014,alshayeji2015detection, berenguel2016banknote} \\
    Image filtering         &  Otsu, median grey-level     & \cite{elkasrawi2014printer} \\
    \cline{2-3}
    \multirow{2}{*}{Brightness normalization}   & brightness map & \cite{bruna2013} \\
                                                & gradient filter & \cite{mandridaketowards} \\
    \cline{2-3}
    \multirow{4}{*}{Colorspace}    & b* (L*a*b) & \cite{roy2015machine} \\
        & Y (YIQ) & \cite{yeh2011employing, Bhavani2014} \\
        & RGB average & \cite{Chambers2015} \\
        & Grayscale & \cite{Glock2009,prasanthi2015indian, dewaele2016forensic,elkasrawi2014printer} \\
        & Grayscale Slicing & \cite{alshayeji2015detection} \\
    \bottomrule
  \end{tabular}
  \end{center}
  \label{tab:preprocessing}
\end{table}

\subsection{Feature extraction}
\label{subsection:approaches_feature_extraction}

Once preprocessed the image, next step is to extract features of interest which could repeat a common pattern in the genuine security documents easily distinguishable against the counterfeit documents. Next we detail some common works center in banknote counterfeit features extraction.

A Free From Deformation (FFD) model for banknote image registration is proposed in~\cite{jin2008hierarchical}, see Table~\ref{tab:feature_extraction}. The authors propose to map a mesh of control points, and measure the deformed position of each pixel by a tensor product of cubic B-splines. Afterwards the map is compared against a reference image with a energy cost function to detect dissimilarities. The authors in~\cite{bruna2013} also use IR images for invisible ink inspection. They base their work in learning the patches, locations and thresholds from the most discriminant IR patterns such that the intra-class distance is minimized, whereas the inter-class distance is maximized. The work is \cite{chae2009study} uses UV information. They calculate similarity with a UV reference image, using a simple pixel sum and remainder comparison. A similar approach is used with the grayscale image in~\cite{prasanthi2015indian}, where the authors after binarize the test image, and they compared the thresholded pixels against a binarized reference image. Several cropped anti-counterfeit regions are check following this procedure, such as microprinting, watermark, security threads, etc. 

The authors in~\cite{yeh2011employing,Bhavani2014}  discard the chrominance information and use only the Y component to build a 256 bin luminance histogram. GLCM features are additionally concatenated with the histogram in \cite{Bhavani2014}. In~\cite{alshayeji2015detection} higher order bit levels are evaluated for grayscale banknote images with the application of Canny edge detection algorithm. They observed that the edges obtained using bit-plane sliced images are more accurate and can be detected faster than obtaining them from the original image without being sliced. A single threshold value-based pattern segmentation method may have difficulty segment the patterns of the UV image. A Gaussian mixture model (GMM) and Expectation maximization (EM) algorithm can be applied to consider the multi-modal characteristics of the UV histogram~\cite{lee2010image}. 
Feature vectors containing the cross-correlation with a synthetic template statistics, the cross-section along the edge and the projection across the edge can be use to distinguish counterfeits made with inkjet and laser printer from genuine bills~\cite{dewaele2016forensic}. Authors in~\cite{roy2015machine} extracts 18 features to check for ink, security thread, Guilloches and Intaglio. They check the denomination region of the banknote which is printed using Intaglio. 9 features are used for this region, such as pixel dominant intensity, hole count, average hue, contrast, etc. For ink, they use colour composition and ink fluidity. Microletter is validated with the spread distribution of the b* component of the L*a*b* color space. The security thread, is composed of two binary features. The first is check for the registration, where the thread should always appear as a single line. The second is determine the presence of text in the thread by pattern matching. Dot distribution, along with cluster distribution and dot density for the strokes, is used as features for Guilloches. Also a latent image, printed with Intaglio is evaluated for inconsistency in the sharpness of the lines, by using a MLP-NN classifier. 

Wavelets appear to be suitable for digital image texture analysis, because they allow analysis of images at various levels of resolution. Authors differentiate genuine and counterfeit Intaglio banknote features by first order statistical moments of wavelet coefficients, using 2D incomplete shift invariant wavelet packet transform (2D-SIWPT)~\cite{Glock2009}. Instead of decomposing the full wavelet packet tree (WPT), the authors also proposed a Best Branch Algorithm (BBA). This algorithm focuses in the branches with highest spatial frequency which contain significant texture characteristics and prunes the rest. Once the best nodes of each scale level are selected an histogram of wavelet coefficients is built. They only used the $\sigma^2$ and kurtosis of the histogram as features to detect counterfeits. Following the previous work, skewness and the local adaptive cumulative histogram (LACH) features of the histogram are added, \cite{Lohweg2012,Lohweg2013}.  The LACH features divide the histogram into three areas: left, middle and right. Being the middle part centered at the zero value coefficient with a width of $\sigma^2$. Afterwards the coefficients in each area are accumulated forming three different score features. The most important lower-frequency DWT coefficients are used in~\cite{Chang2007} for spectral analysis of watermark, OVI and fluorescent fibres. 

Authors in~\cite{berenguel2016banknote,aberenguel:econterfeit} focus on background textures for banknotes and IDs. They represent several ROIs of the background textures as the encoding of dense SIFT descriptors with a learned sparse dictionaries. SIFT-BoW, K-SVD and ScSPM algorithms are used to create the sparse algorithms. Later, same authors in~\cite{aberenguel:evaluation} do a comparison study and statistical evaluation of different state-of-art descriptors to analyze which works best to detect counterfeits when evaluating the background textures. 

Most of the features used for different kinds of text related security documents can also be applied for ID security documents. Authors in~\cite{bertrand2013system} proposes a document forgery detection method based on document’s intrinsic features at character level, such as font properties, character shapes, and character/word alignments. They aim to detect marks such as misalignment or skew found in Scan-Edit and Print (SEP) forged documents. They use a feature vector of Hu moments to detect character similarities, and character size, character horizontal alignment and character principal inertia axis as the feature vector for conception errors. Later, same authors propose to use a conditional random field model which first allows to recognize and classify typefaces, highlighting font forgeries~\cite{bertrand2015conditional}. The CRF model, describes the correlation between fonts, styles, and sizes of the characters. Measuring the probability that a character belongs to a specific font by comparing the font features with a knowledge database, to know whether the character is genuine or fake. Also using a CRF model, the authors in~\cite{satkhozhina2013optical} focus in font recognition, predicting the typeface, weight, slope, and size of the fonts without knowing the content of the text.

It is possible to differentiate between laser and inkjet printing focusing on the edge areas of the printed letters. A feature vector is formed with line edge roughness, area difference, correlation coefficient and texture~\cite{lampert2006printing}. Sharper edges indicates laser printer, meanwhile the opposite correspond to a injket printout. Similarly, authors in~\cite{gebhardt2013document} differentiate laser from inkjet printed pages looking at the edges of the characters for possible forgery attempts. It is possible to differentiate different types of inkjet and laser printers by just looking at the noise produced by the printing technique~\cite{elkasrawi2014printer}. This approach is independent of the document content or size. Authors in~\cite{mandridaketowards,pic2014docscope} use a composite representation based on 27 different criteria (from basic local gradient magnitude to SURF on FFT, or wavelets) to identify the different printing process.  

Text-line alignment and orientation measurement for forgery detection is analyzed in~\cite{van2013text}. They detect implausible skew angles or alignment distances. A considerable area of IDs belongs to non-static data, usually corresponding to the actual content of the document. In~\cite{ahmed2014forgery} propose a framework to detect forgeries only focusing on the static part of non-IDs printed documents. Although the non-static part is also mandatory to check for the final authentication, their approach can be used to determine automatically the variable and non-variable regions of the documents. 

\begin{table}
 \caption{Comparison of counterfeit works based on the security measure and method to extract features.}
  \begin{center}
  \resizebox{\textwidth}{!}{%
  \begin{tabular}{lccc}
    \toprule
    Type & Security Measure & Method & References \\
    \midrule
    Substrate & Document Brightness & Luminance Histogram & \cite{yeh2011employing, Bhavani2014, Desai2014ImplementationOM} \\
    \midrule
    \multirow{4}{*}{Ink} & IR Patterns, Defects & Free From Deformation (FFD) &  \cite{jin2008hierarchical} \\
                        & IR Patterns & Pixel segmentation &  \cite{bruna2013} \\
    \cline{3-4}
                        & \multirow{2}{*}{UV Patterns} & Pixel sum and remainder &  \cite{chae2009study} \\
                        &                               & GMM + EM &  \cite{lee2010image} \\
    \midrule
    \multirow{11}{*}{Printing} & \multirow{2}{*}{MicroPrinting, Guillochés}  & Pixel segmentation &  \cite{bruna2013, prasanthi2015indian} \\
     &   & Canny + Pixel segmentation &  \cite{alshayeji2015detection} \\
    \cline{3-4}
     & \multirow{2}{*}{Intaglio} & 2D-SIWPT + BBA  &  \cite{Glock2009} \\
     &  & 2D-SIWPT + LACH  &  \cite{Lohweg2012,Lohweg2013} \\
    \cline{3-4}
     & \multirow{4}{*}{Printing process} & Edges + Cross Correlation &  \cite{dewaele2016forensic} \\
     &  & edge roughness, area difference, \dots    &  \cite{lampert2006printing} \\
     &  & character edges    &  \cite{gebhardt2013document} \\
     &  & $\mu$, $\sigma$, skewness, kurtosis    &  \cite{elkasrawi2014printer} \\ 
    \cline{3-4}
     & Typography, Intrinsic & CRF, Hu moments, \dots   &  \cite{bertrand2015conditional,bertrand2013system,satkhozhina2013optical} \\
     & Intaglio, Guilloches, Vignettes & texture features  & \cite{berenguel2016banknote,aberenguel:evaluation,aberenguel:econterfeit} \\
     & Guilloches, Printing process & 2nd order statistics, spectral analysis    &  \cite{mandridaketowards,pic2014docscope} \\ 
    \midrule
    Ink,Substrate,Printing & \makecell{Intaglio, Color\\ MicroLetter, Ink Fluidity \\ Security Thread, Guilloches} & \makecell{Dominant intensity, textural similarity \\ colour composition, pixel distribution \\ dot distribution and density, Otsu, \dots}  & \cite{roy2015machine} \\
    Substrate, Ink & OVI, fluorescent fibres, watermark  & DWT & \cite{Chang2007} \\
    \bottomrule
  \end{tabular}
  }
  \end{center}
  \label{tab:feature_extraction}
\end{table}

\subsection{Classification}
\label{subsection:approaches_classification}

The final step in the pipeline of counterfeit authentication is to produce a binary response if the document is either genuine or counterfeit. It is possible to compare feature vectors using Euclidean or Mahalanobis distances~\cite{bertrand2013system}. They compare test characters with a dataset of genuine character. Afterwards a threshold is applied to decide if it is a genuine or fake character. This approach is fast an efficient if the feature vector are enough representative of their classes. 

Most of the works presented use Support Vector Machine (SVM) as the preferred classifier\cite{Bhavani2014,roy2015machine,Glock2009,Chang2007,lampert2006printing,elkasrawi2014printer,berenguel2016banknote}. Linear Discriminant analysis (LDA) is preferred if all elements of classes, follow a gaussian distribution, contribute to the solution uniformly and the possibility of the misclassifying unknown data is fairly low. A multi-stage LDA classifier is used for mobile device banknote counterfeit detection using adaptive wavelets for the analysis of different print patterns on a banknote~\cite{Lohweg2012,Lohweg2013}. 

Choosing a kernel function or hyperparameters in advance for SVM may lead to bad performance. They focus on determine the best kernel  function  and  the associated  kernel hyperparameters. An specific kernel kernel for SVM is associated for each partition of the features in \cite{yeh2011employing}. The combined matrix is calculated with a linear weighted combination of the multiple kernels. Semi-definite programming (SDP) learning is used to obtain optimal weights for the kernel matrices. In~\cite{roy2015machine} they use a combination of two SVM, with Poly and RBF kernel, and an ANN. Then a majority vote approach is followed in integrating results from these classifiers. 

Bayes theorem, describes the probability of an event, based on prior knowledge of conditions that might be related to the event. Bayes can be used for instance to model the conditional probability that two consecutive characters are written with different fonts~\cite{bertrand2015conditional}. In~\cite{aberenguel:evaluation,aberenguel:econterfeit} the authors use SVM to predict a genuine or counterfeit binary label value determining the ROI authenticity. As they check several ROIs, later they learn a Na{\"\i}ve Bayes classifier according to multivariate Bernoulli distributions. Na{\"\i}ve Bayes classifier will penalize the non-occurence of a prediction, meaning they the majority of the ROIs predicted as genuine. 

In~\cite{gebhardt2013document} they use $k$-NN for unsupervised anomaly detection to detect documents printed with a different printing technique than the majority of the documents. This yields the advantage that even unknown printing techniques can be detected. Table~\ref{tab:classification} summarizes the different classifiers used in this section
for counterfeit detection.

\begin{table}
 \caption{Comparison of works based on classifiers.}
  \begin{center}
  \begin{tabular}{cc}
    \toprule
    Method & References  \\
    \midrule
    Homogeneity-based deterioration energy (BDE) & \cite{jin2008hierarchical} \\
    Artificial NN & \cite{roy2015machine,zhao2010study} \\
    SVM & \cite{Bhavani2014,roy2015machine,Glock2009,Chang2007,lampert2006printing,elkasrawi2014printer,berenguel2016banknote} \\
    SVM + Na{\"\i}ve Bayes & \cite{aberenguel:econterfeit,aberenguel:evaluation} \\
    LDA & \cite{Lohweg2012,Lohweg2013, dewaele2016forensic} \\
    Multiple kernel SVM & \cite{yeh2011employing,Desai2014ImplementationOM} \\
    Euclidean, Mahalanobis distance & \cite{bertrand2013system} \\
    Bayes probability & \cite{bertrand2015conditional} \\
    $k$-NN & \cite{gebhardt2013document} \\
    \bottomrule
  \end{tabular}
  \end{center}
  \label{tab:classification}
\end{table}

\subsection{Summary results discussion}
\label{subsection:approaches_summary_results_discussion}

Most of the papers cited in this work report accuracy as the results measure, which is defined as a ratio between the correctly classified samples to the total number of samples. However in counterfeit detection we deal with imbalance datasets, being the accuracy measure clearly misleading for reporting results when there exists a big difference between the positve and negative samples. The word \emph{positive} is used in the sense of counterfeit, whereas the word negative is used in the sense of genuine. False positive rate (FPR), also called false alarm rate (FAR), represents the ratio between the incorrectly classified negative samples to the total number of negative samples. False negative rate (FNR) or miss rate is the proportion of positive samples that were incorrectly classified. Both FPR and FNR are not sensitive to changes in data distributions and hence both metrics can be used with imbalanced data. In Table~\ref{tab:summary_results} we present the results reported by their correspondent authors in the literature. This comparison only includes the works with an ID or banknote counterfeit dataset. We also have not included the works that did not report the results, or the one without enough information of the dataset use for the result. However in the comparison we also talk about non-ID works, which focus in text security documents. Some of this approaches could be transferred to ID security documents.

First we start with banknote related results. In~\cite{Chang2007} claims that lower-frequency DWT coefficients works very effectively and keeps important characteristics for OVI, watermark and fibres feature extraction. The extracted features from~\cite{Glock2009,Lohweg2012,Lohweg2013} allows to separate linearly without error all the elements in the dataset. The work in~\cite{chae2009study} is though for low quality, high-speed inputs for real-time counterfeit detection experiments. So the actual comparison should be improved with a more complex pattern recognition for more complex acquisition scenarios. 

Discarding the deteriorated security document can be a good practice to not raise unnecessary false alarms, if the IR image is available~\cite{jin2008hierarchical}. It can also be used to check the anti-counterfeit IR patterns. Learning the most discriminatory IR patches could be an issue when a highest variability of banknotes is presented in the approach from~\cite{bruna2013}. Although authors in~\cite{yeh2011employing} reports perfect results, they have a small dataset and their method requires more than 13 seconds to compute. Authors in~\cite{Bhavani2014}, do not provide enough information about the configuration of the dataset for the reported results. Moreover, luminance features are not robust under uncontrolled lighting conditions.

Laser and inktjet printers are unable to produce similar sharp edges due to the dithering and satellite droplets respectively. In~\cite{dewaele2016forensic} demonstrate how it is possible to differentiate genuine printed banknotes from inkjet and printed ones using the edge information from mobile phone acquire images. In their studies only the projection along the edge has an acceptable performance in terms of low FPR. The multi-security feature authentication analysis from~\cite{roy2015machine}, can be useful to use some of the low complexity algorithms presented as complementary features for other approaches. The only two classes used for this work are not representative for generalization of the proposed methods to other security documents. 

The approach presented in~\cite{berenguel2016banknote,aberenguel:evaluation,aberenguel:econterfeit} for background texture counterfeit detection in ID and banknotes presents good results evaluating different ROIs of the document. However, several classifiers are trained for each one of the different background texture ROIs. Hence each classifier is treated as an independent problem when in reality they are correlated. This can be considered as an advantage but at the same time a drawback. The advantage is that generating individual models foreach ROI allows to introduce new documents to the system without the need of retraining previous models. The drawback is that trained models do not generalize to other unseen textures. The ideal would be to create a single model able to learn the loose of resolution produced by the scan-printing operation, for all the documents with security background.

Focusing in text related document forgery, the method propose in~\cite{lampert2006printing} also proposes to differentiate laser and printer. However they present an error rate of 5.2\%, which is too high for a fully automatic system. This is cause because a single forged letter makes the whole document a forgery in their method, which is much to sensitive. The approaches in~\cite{bertrand2013system} works with binarized low-resolution documents and do not specify if they included IDs in their datasets, which makes uncertain how it will work under the presence of, for instance kinegrams, occluding the characters. In their later work~\cite{bertrand2015conditional} the same problem of the dataset applies and the performance with banknotes or IDs needs to be tested. The algorithm for text-line alignment and orientation in~\cite{van2013text} only works with pure text documents. This algorithm should be adapted for the presence of images and security features present at IDs and banknotes. Table~\ref{tab:summary_results} summarizes the different classifiers used in this section
for counterfeit detection

\begin{table}
 \caption{Comparison of counterfeit works based on classifiers.}
  \begin{center}
  \begin{tabular}{cccccccc}
    \toprule
    Method & $\mu$FPR & $\mu$FNR & $\mu$Acc & $\mu$Auc & $\mu$F1 & References  \\
    \midrule
    IR Registration similarities & 4.7\% & 7.7\% & - & - & - & \cite{jin2008hierarchical} \\
    IR pixel segmentation & 4.3\% & 0.0\% & - & - & - & \cite{bruna2013} \\
    UV pixel comparison & - & - & 100.0\% & - & - & \cite{chae2009study} \\
    Luminance Histogram + GLCM + SVM  & - & - & 85.0\% & - & - & \cite{Bhavani2014} \\
    Luminance Histogram + Multiple Kernel SVM  & 0.0\% & 0.0\% & 100.0\% & - & - & \cite{yeh2011employing,Desai2014ImplementationOM}\\
    Multiple features + MLP-NN + SVM + ANN  & - & - & 100.0\% & - & - &     \cite{roy2015machine} \\
    2D-SIWPT + BBA + SVM  & 0.0\% & 0.0\% & - & - & - &     \cite{Glock2009} \\
    2D-SIWPT + LACH + LDA  & 0.0\% & 0.0\% & - & - & - &    \cite{Lohweg2012,Lohweg2013} \\
    DWT + SVM  & - & - & 99.0\% & - & - & \cite{Chang2007} \\
    SCSPM + FV + SVM & - & - & - & 0.99 & - &    ~\cite{berenguel2016banknote} \\
    SIFT + FV + SVM + Na{\"\i}ve Bayes & 2.88\% & 2.9\% & - & - & 0.98 &  ~\cite{aberenguel:econterfeit} \\
    Diff. Textures + SVM + Na{\"\i}ve Bayes & - & - & - & - & 0.99 &  ~\cite{aberenguel:evaluation} \\
    \bottomrule
  \end{tabular}
  \end{center}
  \label{tab:summary_results}
\end{table}

\section{Systems and applications}
\label{section:systemsAndapplications}

The previous section~\ref{section:approaches-methodologies} presents different approaches or methodologies to solve the problems of counterfeit detection in banknotes or IDs. At this section we focus in complete systems and applications integrated in smartphone for counterfeit detection. One of the main purposes as explained in section~\ref{section:effects_society_and_document_experts} is to provide tools to security document reviewers and citizens for accessible, affordable and comprehensible counterfeit detection.

Targeting mobile-based solutions for banknote counterfeit detection, authors in~\cite{rahman2017android} use statistical features, and surface roughness of a banknote to represent its properties such as paper material, printing ink, paper quality, and surface roughness, however their dataset is rather small. Researchers in~\cite{dittimi2018mobile} fuse shape context, SIFT, gradient location and orientation histogram (GLOH), and Histogram of Gradient (HOG) into an ensemble base classifier. They focus in banknote recognition, but their system can also detect obvious counterfeits. They achieve real-time on-device processing times, because they only use the quantity number region of the banknote. Authors in~\cite{roy2015machine} propose a system checking multiple anti-counterfeit features, covering the substrate, ink and printing features to create a final decision of the authentication of the document. They also perform a comparison between forensics experts, bank employees and their systems to show its accuracy and time, to deploy their system for mass checking of currency notes in the real world. The authors claim to kept optimal the  complexity of the overall system, to be able to run in a low cost hardware, however they are reporting processing times of 20 minutes. 

In~\cite{aberenguel:econterfeit} the authors build a service-oriented architecture (SOA) end-to-end counterfeit background validation framework. They validate the backgrounds in regions that contains Guilloches, Intaglio or Vignettes. The mobile-server framework provides a low cost solution fit to be used with common smartphones, flexibility for further updates and robust validation methods. The end-to-end system is composed of a Server Framework, a Mobile client and Counterfeit module. The Mobile client tasks are to crop the document on a smartphone, send it to the Server and show the response. The Server uses different technologies to receive the image, store it into a Database and the send a query to the Counterfeit module to validate the image. The Counterfeit module connect to different frameworks to validate the background textures of the input image. They report that the complete process of acquiring and receiving the results for the user, sending an image at 400 dpi is of 9.2s. However, the validation of the background texture alone is only 620ms.

Focusing on the Intaglio regions, the authors in~\cite{Lohweg2012,Lohweg2013} propose a system of counterfeit detection able to run on smartphones. They propose to use Y-channel, from the preview image in YCrCb format due it seems to be uncompressed. This way they avoid the JPEG-artifacts distortions, which affects the high frequency components. Also this channel from the preview image has low resolution, therefore usable for image processing applications. The preview image with $1280\times720$ is used, and a pre-processing part of the camera module generates in one step a grey-scale image in the preferred size of $400\times400$ pixel for the top, middle and bottom region, with an approximate resolution of 620 dpi. The Intaglio zones have been also validated by the authors in~\cite{pfeifer2016detection}, a two-dimensional discrete Fourier transform (2D DFT) is use to identify the periodical screens within the separated channels. Afterwards, unimportant
frequencies are suppressed, transforming back the remaining ones. Finally, a model-based feature extraction determines the screen angles and the individual offset rosettes. They acquire their dataset with a minimum of 1100 dpi within a controlled environment. 

The applications of fingerprints in ID documents is discussed in~\cite{yang2014fingerprint}. They propose a enrollment and verification system, that ultimately matches the collected fingerprints with the ones stored in a chip integrated on the ID. These system faces problems in case of wet, dry, or scratched fingerprint. Authors in ~\cite{schouten2009biometrics} discusses the requirements, design and application scenario’s of multi-modal biometrical systems. They study the privacy, security at biometric passports and the public perception and repudiation of these security measures. Other researchers have focused in a banknote validation method which uses radio frequency identification (RFID) and an NFC-enabled smartphone for real-time banknote validation~\cite{eldefrawy2015banknote}. Although this method is computationally less expensive than other methods not all the documents and banknotes have the RFID chips. Ordered by year, Table~\ref{tab:systems_apps} presents the different works mentioned at this section.

\begin{table}[h]
 \caption{Comparison of systems and applications.}
  \begin{center}
  \begin{tabular}{ccccc}
    \toprule
    Year & Type & Security Measures & Processing time & References  \\
    \midrule
    2018 & \multirow{4}{*}{Smartphone App.} & Quantity number & 0.02ms & \cite{dittimi2018mobile} \\
    2017 &  & Intaglio, Guilloches, Vignettes  & 9.2s & \cite{aberenguel:econterfeit} \\
    2017 &  & Substrate quality/roughness, ink & - & \cite{rahman2017android} \\
    2016 &  & Intaglio & - & \cite{pfeifer2016detection} \\
    \cline{2-5}
    2015 & System & substrate, ink, printing & 20m & \cite{roy2015machine} \\
    \cline{2-5}
    2015 & \multirow{2}{*}{Smartphone App.} & Intaglio & 1.2s & \cite{Lohweg2012,Lohweg2013} \\
    2015 &  & RFID chip, NFC-enabled & real-time & \cite{eldefrawy2015banknote} \\
    \cline{2-5}
    2014 & \multirow{2}{*}{System} & Fingerprint & - & \cite{yang2014fingerprint} \\
    2009 &  & multi-modal biometrics & - & \cite{schouten2009biometrics} \\
    \bottomrule
  \end{tabular}
  \end{center}
  \label{tab:systems_apps}
\end{table}

\section{Trends}
\label{section:trends}

For centuries legitimate authorities and equally determined fraudsters have fought in an attempt to enhance or copy the technology of security documents. The good news is there are fewer counterfeit documents than there used to be. The bad news is counterfeiters are getting better at their craft. To solve the increasing difficulty to detect the high quality counterfeits, it is gaining popularity the idea of removing from circulation the physical security documents and move towards a digital world.

In ~\cite{wolman2013end} the authors present from the invention of physical money to the evolution of paperless alternatives. They also present some insights of a "cash-less society" as either a big pro of a big con, doing an intriguing survey of what will happen to counterfeiters and others in the coming cashless society. Countries are eliminating cash at varying speeds. Reports shows that four out of five purchases in Sweden are paid for electronically or by card~\cite{sweden_cashless}, there exists predictions which consider Sweden will be the first country in the world completely free of cash. Another contender for the first cashless country in the world is in China, embraced with the QR codes~\cite{nayax_first_cashless_country}. Other cases, like UK, does not fall behind the cash-less future, where credit/debit cards, contact-less and online payments have taken over cash payments. Smartphone applications transforms the device into a seamless wallet, remittance, and payments tool, granting financial inclusion and unprecedented convenience to billions of unbanked people around the world. Furthermore the rise of decentralised cryptocurrencies is starting to coexist legitimately alongside digital currencies. Removing physical cash, automatically will remove counterfeit banknotes, consequently large-scale criminal activity would be much easier to detect: transactions will have to bypass bank accounts, which are traceable. 

Similarly, ID security documents are moving already to Electronic national ID cards (e-IDS). e-IDs include a microprocessor for stronger document verification but also on-line authentication and signature. These e-ID cards offers the best identity theft protection and also enable governments to implement on-line applications such as eGovernment solutions, giving citizens access to public services with the reassurance of robust security. According to~\cite{acuity_eID_report}, the number of electronic National ID cards in circulation will reach 3.6 billion citizens by 2021. Furthermore, another strong and trusted method of identification, is the mobile ID (mID), which are mechanisms using an eID component for accessing online services via mobile devices~\cite{gemalto_trends}. Pioneers countries, like Austria, Estonia, Finland, Norway, and Turkey are moving towards mID. Moreover, several US states have launched pilots for digital driver licenses, also called mobile drivers license. It provides an on-screen mobile version of the traditional photo and driver information, being highly secure and with stronger counterfeiting characteristics. Being able to update instantly the driver data information and facilitates real-time communication.  

Does this means researchers should stop improving anti-counterfeit features for the security documents? Quite the opposite. Physical IDs and banknotes are not going to disappear in a near future. Today, $85\%$ of worldwide consumer spending is done in cash despite many forecasting the demise of this resilient product. The conclusion that the near and midterm future of cash is not a non-cash society, but the one with less cash~\cite{watermark_conference_2017}. Printing, circulation, and reprinting of banknotes and the growth in tourism is projected to increase the demand for passports and visas, which in turn, is driving the growth of the security paper market across the globe. The security paper market is projected to grow from USD 11.4 billion in 2018 to USD 14.8 billion by 2023~\cite{business_wire}. Hence counterfeit IDs and banknotes will continue circulating. 

It has been observed that after the introduction of polymer substrate paper banknotes, the average quality of the polymer substrate counterfeits has also increased. Over the past two years, around 40 per cent of counterfeits detected in Australia have been considered high quality~\cite{reserve_bank_of_australia_trends}. Authors in~\cite{keller2018inkjet} created a way to print chromatic holograms on any surface and also create high-quality organic piezoelectric structures. Printing luminescent structures based on \emph{nanoparticle ink} allows for the fabrication of custom holograms by means of a common inkjet printer, which can produce anti-counterfeiting objects with high stability and durability. Private companies are also developing new security features, like \emph{Galaxy Threads} or \emph{RollingStar Threads}~\cite{G_D_Currency_Technology}. The former produces three optically variable effects combined in this thread: 3D motifs, dynamic effects and color shift, are visible from any perspective. The later makes possible dynamic effects that immediately attracts attention, the thread links motion sequences and color shift. \emph{Satellite holograms} and \emph{Emerald number}  have been recently introduced in the \EUR{100} and \EUR{200} banknotes~\cite{european_central_bank_100_200}. The first makes and observable \EUR{} symbols moving around the number when tilting the banknote and become clearer under direct light. The second is an OVI that changes the colour from emerald green to deep blue of the number, producing an effect of light that moves up and down when tilting the banknote.

Most countries are constantly adding new security features or upgrading the existent ones at each of its banknotes and IDs. The aim is to make harder to counterfeit, but still easy to check. In this section we have just cited a few of the actual improvements and the current and future trends.

\section{Conclusion}
\label{section:conclusion}

We present in this work a survey of identity document and banknote security forensics. We focus in the anti-counterfeit security measures which can be solved automatically by computer vision algorithms. Initially we compare this work with multiple state-of-art surveys to show the completeness and the need of writing a new survey. We add sections which were not usually treated at other surveys, such as history, effects of forgery in society or document experts, with the objective that readers without the previous knowledge in counterfeit detection understand the basics and difficulties of this field. The sections anti-counterfeit measures explains how are designed the most known anti-counterfeit features based in three categories: The security substrates, the security inks and the security printings. With this categories we cover all the stages of production of a security document. We add also in these section what are the possible attacks a counterfeiter usually performs to forge an ID or a banknote. 

To our knowledge we present in this work on of the most completes studies in anti-counterfeit features for security documents. One of the purposes of this work was to provide a reference for future researchers of the available datasets that they can work to continue developing new algorithms and techniques for counterfeit detection. Unfortunately the availability of ID and banknote datasets for this field continues to be an issue. Further study should be done in how it could be possible to create these datasets, without infrincting PII and copyright issues, to share for the research community and create a baseline of results. 

In this work, we focus only in identity and banknotes. Both share similar anti-counterfeiting security features. A future line of work, would be doing the same analysis of tamper-evident labels, cheques, product authentication, stock certificates, postage stamps, etc. Then compare which of these objects has similar security features and which ones has easier available datasets. The objective would be two-fold. First study to possibility of using security detection algorithms of one object to the other. Second study if its possible to transfer knowledge between models trained in one printing security object dataset to the other. An objective of this work was to link each presented security feature with its corresponding research. The algorithm in each research publication, should preferably be designed with computer vision algorithms. This objective has not been entirely successful, due some of the presented security features it not clear there is a published research for them. 

At the approaches section we present the state-of-art approaches for counterfeit detection on IDs and banknotes. We categorize this section into preprocessing, feature extraction and classification. Ultimately we discuss some of these works and the pros and cons of applying them for future research. Here we believe that future works on counterfeit detection, should be clear about the dataset information they used to generate the results. They should also use FPR and FNR metrics which are much more representative than accuracy metrics. Both FPR and FNR are not sensitive to changes in data distributions and hence both metrics can be used with imbalanced data. 

From the presented anti-counterfeit measures at this work, we conclude that today, there no exists a single deterrent feature which is readily recognizable, highly durable, difficult to counterfeit or simulate, costly affordable, and easy to produce. A best strategy is to select a combination of features, which adds complexity to the counterfeiter's task and increase the number of counterfeiting steps to the point that the casual counterfeiter would eventually "give up". Discouraging the counterfeiter is harder and is best accomplished by having a larger number of anti-counterfeit features, each requiring a different means and material for simulation. The objective here is one of attrition, overwhelming the counterfeiter with so many tasks. Stopping dedicated professional counterfeiters is very difficult, the only thing it can be done is to periodically add new features designed to produce delays into the counterfeiter's production cycle. Same reasoning can be applied to design computer vision algorithms for counterfeit detection. The strategy of combining algorithms targeting different anti-counterfeit measures is the best strategy to follow for a robust detectors.

\section{Acknowledgment}
This work has been partially supported by the Spanish Research Project RTI2018-095645-B-C21 and the Industrial Doctorate Grant 2014 DI 078 with the support of the Secretariat for Universities of the Ministry of Business and Knowledge and the CERCA Program of the Government of Catalonia.

\bibliographystyle{unsrt}  
\bibliography{ms}  

\end{document}